\documentclass[11pt]{article}

\usepackage[final]{acl}

 \usepackage{microtype}
\usepackage{times}
\usepackage{latexsym}
\usepackage{amsmath}
\usepackage{enumitem}
\usepackage{amssymb} 
\usepackage{mathrsfs}
\usepackage{booktabs}
\usepackage{multirow} 
\usepackage{colortbl}
\usepackage{makecell}  
\usepackage{bbding}  
\usepackage{bbm}
\usepackage{graphicx}
\usepackage[ruled,linesnumbered]{algorithm2e}
\usepackage{algorithmic}
\usepackage{wrapfig} 
\usepackage{subcaption}
\usepackage{adjustbox}
\usepackage{pifont}
\usepackage{xcolor}
\usepackage{tabularx}
\usepackage{tocbibind}
\usepackage{iftex}
\usepackage[most]{tcolorbox}

\usepackage[table]{xcolor}
\usepackage{array}

\usepackage{hyperref}

\ifPDFTeX
  \usepackage[utf8]{inputenc}
  \usepackage[T1]{fontenc}
  \usepackage{CJKutf8}
\else
  \ifLuaTeX
    \usepackage{luatexja-fontspec}
    \setmainjfont{FandolSong-Regular}
  \fi
  \ifXeTeX
    \usepackage{xeCJK}
  \fi
\fi

\title{
  \raisebox{-5.5pt}{%          % ← 上下微调（负值 = 下移）
    \includegraphics[width=0.047\textwidth]{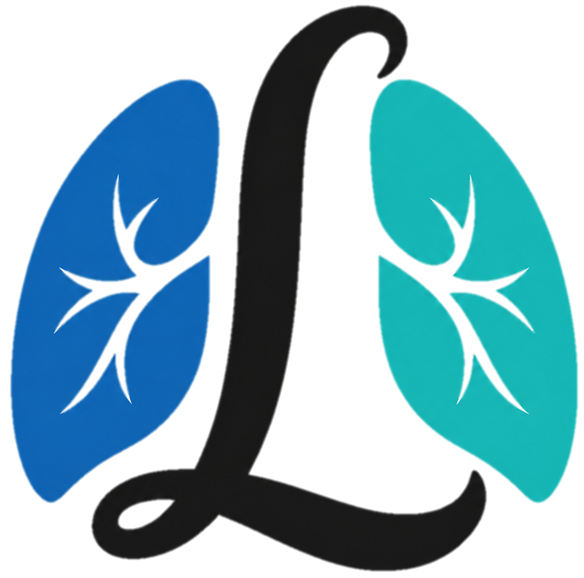}%
  }
  \hspace{-6pt}
  \textit{ung-R1}: A Knowledge Graph-Guided LLM for Pulmonary \\ Diagnostic Reasoning
}

\author{
\textbf{Haoyang Zeng\textsuperscript{1*}, Yuanxi Fu\textsuperscript{1*}, Rongzhen Li\textsuperscript{2*}, Yuming Yang\textsuperscript{1}, Xiao Sun\textsuperscript{1}, Jingwang Huang\textsuperscript{1}}\\
\textbf{Gujie Shao\textsuperscript{1}, Guohui Xiang\textsuperscript{2}, Quan Lu\textsuperscript{2}, Dongfan Ye\textsuperscript{3}, Xuetao Chen\textsuperscript{3}, Jiang Zhong\textsuperscript{1†}, Kaiwen Wei\textsuperscript{1†}, Zhi Xu\textsuperscript{3†}}\\[6pt]
\textsuperscript{1} School of Computer Science, Chongqing University, Chongqing, China \\
\textsuperscript{2} AI Research Institution, Mashang Financial Institution \\
\textsuperscript{3} Department of Information, Third Military Medical University \\
\text{\href{mailto:zenghy@stu.cqu.edu.cn}{zenghy@stu.cqu.edu.cn},
\href{mailto:jiangzhong@cqu.edu.cn}{jiangzhong@cqu.edu.cn},
\href{mailto:weikaiwen@cqu.edu.cn}{weikaiwen@cqu.edu.cn},
\href{mailto:xuzhihxk@tmmu.edu.cn}{xuzhihxk@tmmu.edu.cn}}
}

\begin{document}
\ifPDFTeX
\begin{CJK*}{UTF8}{gbsn}
\fi
\maketitle

\begin{abstract}
Diagnosing pulmonary diseases requires integrating heterogeneous evidence amid phenotypic variability and cross-disease overlap. Although large language models (LLMs) have shown progress on pulmonary knowledge question answering (QA) and information-processing tasks, reliable pulmonary diagnosis requires patient-specific, relation-aware reasoning over electronic medical record (EMR) evidence rather than isolated knowledge recall. We define this gap between pulmonary knowledge and case-level diagnostic reasoning as the \textit{Pulmonary Knowledge-to-Diagnosis Gap}. To address it, we introduce \textbf{LungKG}, the first structured pulmonary knowledge graph for diagnostic knowledge organization and record-grounded reasoning. LungKG contains 59,038 nodes and 164,308 edges across 15 entity types and 112 relation types, serving as both a reusable pulmonary knowledge resource and the foundation for LungKG-guided model adaptation. Built on LungKG, we propose \textbf{Lung-R1}, a LungKG-guided pulmonary LLM trained through KG-constrained reasoning-chain construction and KG-guided reinforcement learning. In a 20-system evaluation, Lung-R1-14B achieves state-of-the-art performance across Choice, Pulmonary-QA, and EMR Diagnosis, reaching an EMR Diagnosis score of 4.3583 and surpassing the strongest non-Lung-R1 baseline by 0.1476 points. These results demonstrate the value of LungKG-guided training for EMR-based pulmonary diagnosis.
\end{abstract}

\begin{figure}[t]
  \centering
  \includegraphics[width=\columnwidth]{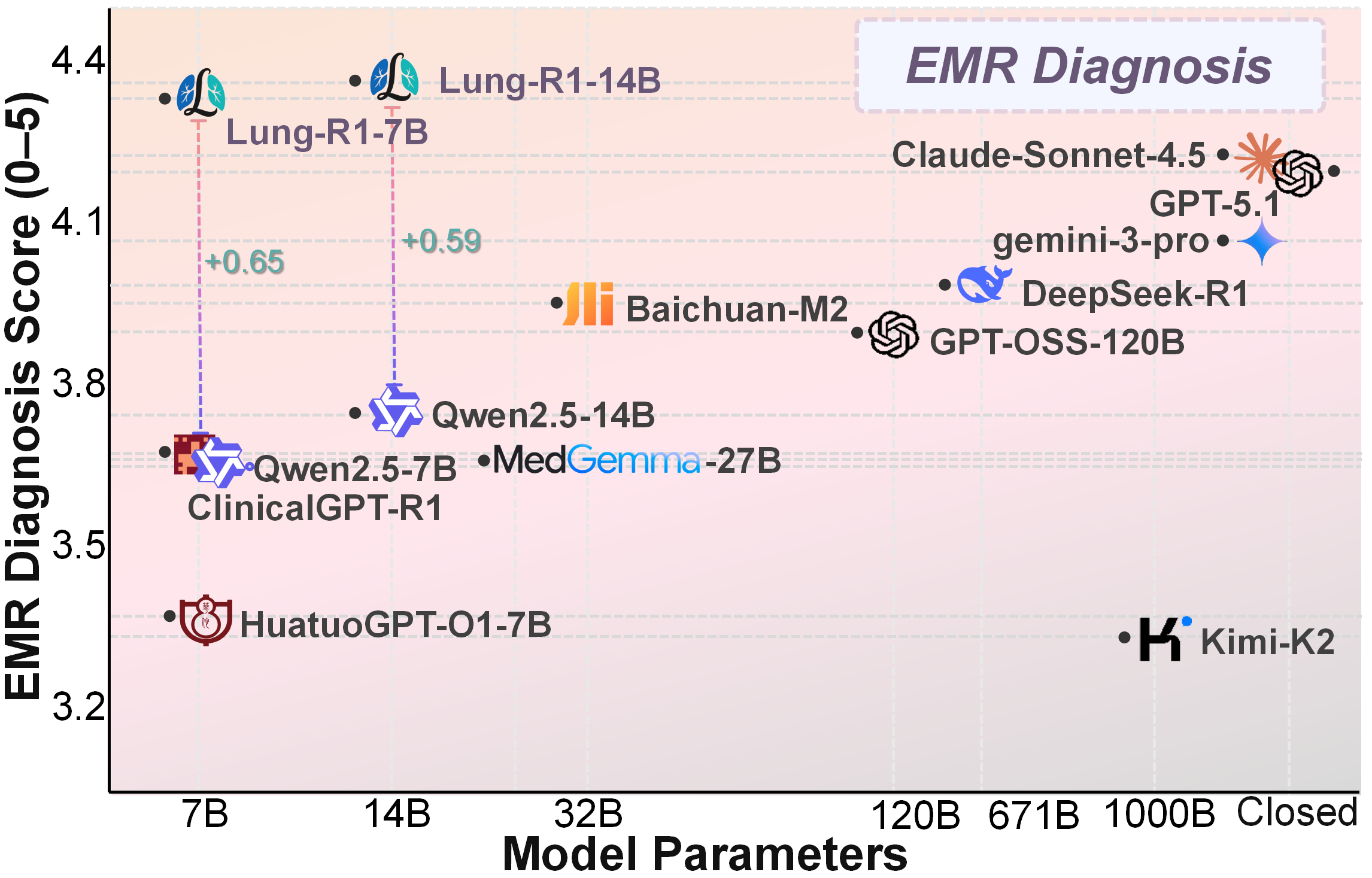}
  \caption{EMR diagnosis performance on the EMR Diagnosis task. Lung-R1 achieves state-of-the-art performance at 7B/14B scale.}
  \label{fig:emr_diagnosis_results}
\end{figure}

\section{Introduction}

Diagnosing pulmonary diseases requires integrating heterogeneous clinical, radiological, functional, and pathological evidence~\citep{raghu2022idiopathic,ryerson2025update,delaney2024meta,khor2023treatable}.
This task is further complicated by the marked phenotypic variability of pulmonary diseases and the cross-disease overlap of clinical and radiological manifestations\footnote{\href{https://www.who.int/health-topics/chronic-respiratory-diseases}{WHO: Chronic Respiratory Diseases}.}~\citep{bender2024global}.
These characteristics make pulmonary diagnosis a demanding setting for evaluating whether large language models (LLMs) can support clinical reasoning~\citep{ahsan2024retrieving}. Existing pulmonary AI resources have advanced pulmonary clinical intelligence across knowledge question answering, radiology understanding, multimodal imaging, and information extraction~\citep{bae2023ehrxqa,rocha2019respiratorysound}.
However, pulmonary diagnosis requires patient-specific reasoning that maps observed evidence to disease hypotheses through clinically meaningful diagnostic relations. Thus, strong performance on pulmonary knowledge and information-processing tasks does not by itself establish the capacity for reliable patient-specific diagnostic reasoning.

We refer to this mismatch as the \textit{Pulmonary Knowledge-to-Diagnosis Gap}: the gap between knowledge-centric medical training and case-centric pulmonary diagnosis. In such training settings, pulmonary facts are usually presented as examination questions, QA pairs, or isolated textual instructions; in clinical practice, the same knowledge must be applied as interdependent evidence chains over a specific patient record. As a result, strong performance on pulmonary knowledge tasks does not necessarily indicate reliable record-grounded diagnostic reasoning~\citep{wang2025knowledge,gao2023large,chandak2023building}.

To address this challenge, we construct \textbf{LungKG}, a structured pulmonary knowledge graph explicitly designed for pulmonary diagnosis-oriented reasoning and model adaptation. LungKG contains 59,038 nodes and 164,308 edges across 15 entity types and 112 relation types, organizing pulmonary diseases, symptoms, pathogens, examinations, imaging findings, drugs, treatments, and diagnosis-related evidence into typed, directed relations. LungKG makes pulmonary diagnostic relations explicit and serves as a reusable knowledge resource as well as the structural foundation for LungKG-guided model adaptation.

Moreover, considering that existing LLMs lack a mechanism to apply pulmonary relations to patient-specific electronic medical record (EMR) evidence, we introduce \textbf{Lung-R1}, a LungKG-guided pulmonary LLM family that transforms structured pulmonary knowledge into diagnosis-oriented reasoning over patient-specific EMR evidence. Lung-R1 is trained through a two-stage framework. \textit{First}, during supervised fine-tuning, \textit{KG-Constrained Reasoning Chain Construction} uses LungKG to generate graph-grounded Chain-of-Thought supervision, which is combined with EMR diagnosis supervision to connect structured pulmonary knowledge with realistic clinical language and record-to-diagnosis targets. \textit{Second}, in \textit{KG-Guided Reinforcement Learning}, KG-guided rewards align model outputs with diagnosis correctness, graph faithfulness, and relation/path consistency.

To evaluate Lung-R1, we construct a pulmonary evaluation suite covering Choice, Pulmonary-QA, and EMR Diagnosis, and benchmark 20 systems. Unless otherwise specified, the main Lung-R1 result refers to the CoT model trained with the full knowledge-graph question answering (KGQA) and EMR data using the two-stage SFT plus KG-guided RL pipeline. It achieves the strongest performance on the three primary metrics: 67.60\% Choice accuracy, 4.416 Pulmonary-QA score, and 4.3583 EMR Diagnosis score. As shown in Figure~\ref{fig:emr_diagnosis_results}, its EMR Diagnosis score exceeds the strongest non-Lung-R1 baseline, Claude-Sonnet-4.5, by 0.1476 points. Our contributions are summarized as follows:

(1) We construct \textbf{LungKG}, the first structured pulmonary knowledge graph for diagnosis-oriented pulmonary reasoning. It contains 59,038 nodes, 164,308 edges, 15 entity types, and 112 relation types, organizing key pulmonary entities and diagnostic evidence into typed, directed relations.

(2) We develop \textbf{Lung-R1}, a LungKG-guided pulmonary LLM trained with KG-constrained CoT supervision and KG-guided reinforcement learning to align diagnostic correctness, graph faithfulness, and relation/path consistency.

(3) We build a held-out pulmonary evaluation suite covering Choice, Pulmonary-QA, and EMR Diagnosis. A 20-system comparison shows that Lung-R1 achieves state-of-the-art primary-metric performance, especially on record-grounded pulmonary diagnosis.

\begin{figure*}[t]
  \centering
  \includegraphics[width=\textwidth]{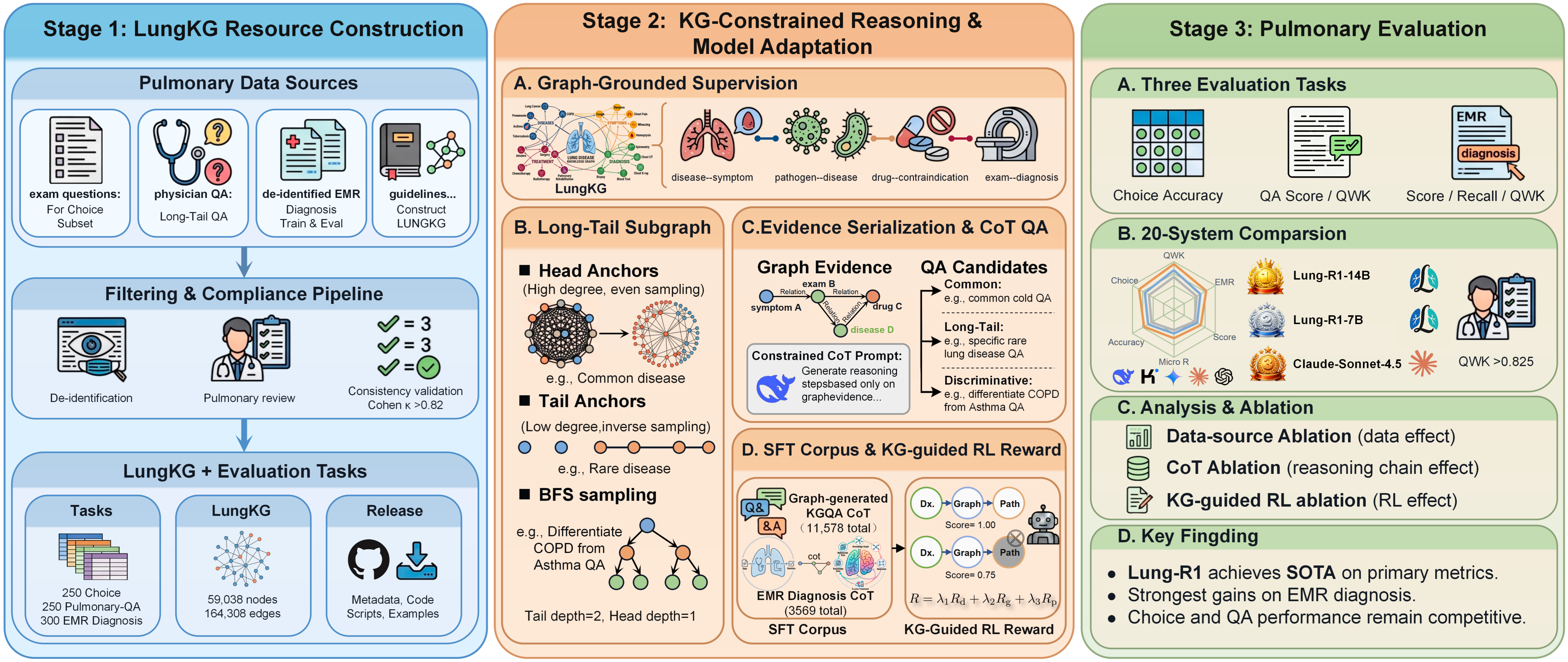}
\caption{
Overview of the LungKG-guided Lung-R1 pipeline: 
(a) LungKG construction from validated pulmonary sources; 
(b) KG-constrained CoT construction, SFT, and KG-guided RL for Lung-R1 adaptation; 
(c) pulmonary evaluation on Choice, Pulmonary-QA, and EMR Diagnosis tasks across 20 systems.
}
  \label{fig:lungr1_pipeline}
\end{figure*}

\section{Related Work}
\textbf{Pulmonary AI resources and medical LLM evaluation.}
Existing pulmonary AI resources mainly focus on specific evidence types or tasks, including chest imaging interpretation, respiratory sound analysis, and multimodal EHR reasoning with radiology evidence~\citep{bae2023ehrxqa,rocha2019respiratorysound}. They capture important pulmonary signals, but do not organize pathogen evidence, imaging/examination findings, contraindications, treatment risks, and patient-specific context into relation-aware supervision for record-grounded diagnosis. Medical LLM benchmarks, including MedQA~\citep{jin2021disease}, MedMCQA~\citep{pal2022medmcqa}, PubMedQA~\citep{jin-etal-2019-pubmedqa}, MMLU-health~\citep{hendrycks2021measuring}, and CMB~\citep{wang2024cmb}, mainly evaluate examination-style knowledge or biomedical QA, while clinical evaluations address realistic tasks and differential diagnosis~\citep{arora2025healthbench,bedi2026medhelm,hager2024clinicaldecision,mcduff2025differential}. EHR resources such as MIMIC-IV~\citep{johnson2023mimic} provide authentic clinical language but lack explicit pulmonary diagnostic relations for graph-grounded supervision and KG-guided rewards.

\paragraph{Knowledge-grounded medical adaptation and alignment.}
Medical language models have progressed from domain-adaptive pretraining to instruction tuning, including BioBERT~\citep{lee2020biobert}, ClinicalBERT~\citep{alsentzer-etal-2019-publicly}, Med-PaLM~\citep{singhal2023large}, PMC-LLaMA~\citep{wu2023pmc}, and MEDITRON~\citep{chen2023meditron}. Retrieval-Augmented Generation (RAG)~\citep{lewis2020retrieval} and GraphRAG~\citep{edge2024local,soman2024biomedical,zuo2025kg4diagnosis} improve factual grounding through inference-time retrieval, while Chain-of-Thought prompting~\citep{wei2022chain}, STaR~\citep{zelikman2022star}, and Self-Instruct~\citep{wang2023selfinstruct} generate or refine reasoning supervision. However, these approaches do not use structured pulmonary relations as training-time supervision and reward-alignment signals. Lung-R1 instead uses \textbf{LungKG} as a diagnosis-oriented substrate for graph-grounded CoT SFT and KG-guided reinforcement learning, shifting pulmonary knowledge grounding from inference-time access to training-time reasoning alignment.

\section{LungKG}
\label{sec:lungkg}

We introduce \textbf{LungKG}, the first structured pulmonary knowledge graph designed for pulmonary diagnosis-oriented reasoning and model adaptation. As shown in Figure~\ref{fig:lungr1_pipeline}(a), LungKG targets the \textit{Pulmonary Knowledge-to-Diagnosis Gap} by reorganizing isolated pulmonary facts into typed, directed relations for diagnosis-oriented reasoning~\citep{anuyah2025automated,zhou2026fine,abu2023healthcare}. Conventional QA supervision presents medical knowledge as isolated question--answer pairs, whereas pulmonary diagnosis requires connecting distributed evidence, including symptoms, imaging and examination findings, pathogen clues, contraindications, treatments, and patient-specific context, into coherent reasoning paths. LungKG therefore serves as an intermediate substrate between pulmonary knowledge resources and Lung-R1 adaptation, as shown in Figure~\ref{fig:lungr1_pipeline}(b).

We build LungKG from diverse pulmonary resources including curated knowledge, clinical guidelines and examination materials. Its standard and reliable construction is ensured by our annotation team and review protocol (Appendix~\ref{app:annotation_team}). We use DeepSeek-R1~\citep{deepseekai2025r1} for entity/relation extraction, normalization and candidate graph building. Entities are unified into standard pulmonary concepts to resolve duplicates from synonyms, abbreviations and textual variations. Relations are converted to typed directed edges such as disease–symptom, pathogen–infection, examination–diagnosis, drug–contraindication and treatment–condition. Clinician-led quality control, deduplication, relation mapping and consistency checks remove noisy evidence. The annotation consistency assessment yields F-agreement of 88.2\% for named entity recognition and 83.8\% for entity relation annotation, supporting the reliability of LungKG construction. The agreement definitions and matching rules are provided in Appendix~\ref{app:lungkg_agreement}, with additional construction details in Appendix~\ref{app:lungkg_entities}. The final LungKG contains 59,038 nodes and 164,308 edges, covering 15 entity types and 112 relation types.

Table~\ref{tab:lungkg_statistics} summarizes the scale, entity inventory, diagnostic relation coverage, and downstream training role of LungKG. LungKG is not intended to be a complete ontology of all medical knowledge; it is a pulmonary-domain substrate for graph-grounded supervision and KG-guided reward construction.

\begin{table}[t]
\centering
\small
\setlength{\tabcolsep}{3pt}
\renewcommand{\arraystretch}{1.06}
\begin{tabularx}{\columnwidth}{@{}>{\raggedright\arraybackslash}p{0.34\columnwidth}>{\raggedleft\arraybackslash}p{0.12\columnwidth}>{\raggedright\arraybackslash}X>{\raggedleft\arraybackslash}p{0.12\columnwidth}@{}}
\toprule
\multicolumn{4}{@{}l}{\textbf{Graph scale and diagnostic role}} \\
\midrule
\textbf{Aspect} & \multicolumn{3}{@{}>{\raggedright\arraybackslash}p{0.63\columnwidth}@{}}{\textbf{Content / role}} \\
Scale & \multicolumn{3}{@{}>{\raggedright\arraybackslash}p{0.63\columnwidth}@{}}{59,038 nodes; 164,308 edges} \\
Schema & \multicolumn{3}{@{}>{\raggedright\arraybackslash}p{0.63\columnwidth}@{}}{15 entity types; 112 relation types} \\
Diagnostic relations & \multicolumn{3}{@{}>{\raggedright\arraybackslash}p{0.63\columnwidth}@{}}{disease--symptom; pathogen--infection; examination/imaging--diagnosis; drug--contraindication; treatment--condition} \\
Training role & \multicolumn{3}{@{}>{\raggedright\arraybackslash}p{0.63\columnwidth}@{}}{KG-constrained CoT construction for SFT; KG-guided rewards for RL} \\
\midrule
\multicolumn{4}{@{}l}{\textbf{Entity-type inventory}} \\
\midrule
\textbf{Entity Type} & \textbf{\#Nodes} & \textbf{Entity Type} & \textbf{\#Nodes} \\
Western medicine & 11,544 & Pathogen & 2,804 \\
Other treatment & 10,976 & Guideline recommendation & 1,928 \\
Disease & 8,448 & Patent medicine & 1,446 \\
Drug & 6,481 & \makecell[l]{Diagnosis/treatment\\technology/equipment} & 1,158 \\
Symptom & 5,120 & Anatomical site & 672 \\
Examination & 4,098 & Clinical department & 538 \\
Epidemiology & 3,090 & Herbal medicine & 400 \\
Surgical treatment & 335 & & \\
\bottomrule
\end{tabularx}
\caption{Core statistics and entity inventory of LungKG.}
\label{tab:lungkg_statistics}
\end{table}

LungKG emphasizes relation types that are central to pulmonary diagnosis: 
(1) pathogen relations, which connect pathogens with infections, diseases, risk factors, and treatment implications; 
(2) symptom and differential-diagnosis relations, which link common presentations such as cough, dyspnea, fever, hypoxemia, and chest tightness to candidate diseases and distinguishing findings; 
(3) imaging and examination relations, which connect chest imaging, pulmonary function tests, laboratory indicators, and examination findings with diagnostic hypotheses; and 
(4) drug, contraindication, and treatment-risk relations, which support reasoning about medication exposure, adverse effects, and patient-specific risks.

These relation groups serve as interface between LungKG and Lung-R1. As shown in Figure~\ref {fig:lungr1_pipeline}(b), LungKG relations are sampled to build KG-constrained Chain-of-Thought supervision and define KG-guided reward signals for reinforcement learning. Detailed reasoning-chain construction and reward design appear in Section~\ref {sec:lung_r1}.

\section{Lung-R1}
\label{sec:lung_r1}

Motivated by the \textit{Pulmonary Knowledge-to-Diagnosis Gap}, as shown in Figure~\ref{fig:lungr1_pipeline}(b), we introduce \textbf{Lung-R1}, a LungKG-guided pulmonary LLM that transforms structured pulmonary knowledge into record-grounded diagnostic reasoning.

\subsection{Training Setup}

We use Qwen2.5-7B-Instruct and Qwen2.5-14B-Instruct~\citep{qwen2025qwen25} as the base backbones for Lung-R1 and adapt them with LungKG-guided training. The training process consists of two stages: supervised fine-tuning (SFT) with LungKG-grounded reasoning data, followed by KG-guided reinforcement learning (RL) to align model outputs with diagnosis correctness, graph faithfulness, and relation/path consistency. Detailed optimization settings and SFT hyperparameters are provided in Appendix~\ref{app:training_config}, and implementation-level RL setup details are provided in Appendix~\ref{app:kg_guided_rl_details}.

\textbf{Training corpus.}
We use two complementary reasoning-supervision sources. The first is knowledge-graph question answering (KGQA) data with CoT reasoning, generated from LungKG subgraphs. Sampled LungKG evidence is converted into reasoning-augmented QA samples under graph-faithfulness and medical-safety constraints. After filtering, duplicate control, and LLaMA-Factory conversion, the final KGQA manifest contains 11,578 samples, each including question, reasoning, answer, and graph evidence.

The second source is a de-identified EMR diagnosis corpus provided by a collaborating clinical institution, containing 3,569 pulmonary diagnosis samples. Each sample includes structured clinical fields, such as chief complaint, history of present illness, physical examination, and laboratory/imaging findings, together with the corresponding pulmonary-related diagnosis. These EMR samples are strictly non-overlapping with EMR Diagnosis evaluation cases. Under LungKG constraints, EMR evidence is linked to graph entities, relations, and paths toward the reference diagnosis, then verbalized as natural-language diagnostic CoT supervision.

The final SFT corpus contains 15,147 reasoning-supervision samples combining KGQA and LungKG-constrained EMR diagnosis data. KGQA generation and EMR reasoning-chain construction are performed by DeepSeek-R1~\citep{deepseekai2025r1} under clinician-guided constraints and post-generation quality control. Filtering prompts are provided in Appendix~\ref{app:prompt_templates} and are used for training-data quality control rather than benchmark judging.

\subsection{KG-Constrained Reasoning Chain Construction}

\textbf{Graph-grounded KGQA construction.}
For KGQA data, we construct reasoning chains from sampled LungKG subgraphs. Given a pulmonary subgraph $g \subseteq \mathrm{LungKG}$, typed nodes and directed labeled edges are serialized into a graph-evidence context, denoted as $\operatorname{Serialize}(g)$. A prompt-based generation procedure then produces a question, reasoning chain, and answer grounded only in the serialized graph evidence:
\begin{equation}
(q, r, a)=\operatorname{Gen}_{\mathrm{KGQA}}(\operatorname{Serialize}(g)),
\end{equation}
where $q$, $r$, and $a$ denote the generated question, KG-grounded reasoning chain, and answer, respectively. $\operatorname{Gen}_{\mathrm{KGQA}}(.)$ denotes the automatic KGQA generation procedure described in Appendix~\ref{app:prompt_templates}. The procedure preserves relation direction, avoids unsupported diagnostic or treatment claims, and expresses reasoning in natural language rather than graph-traversal traces.

To improve coverage of underrepresented pulmonary knowledge, we use reverse tail-degree sampling, which assigns higher sampling probability to low-degree graph regions. Sampled subgraphs are expanded into local communities and converted into common QA, long-tail QA, and discriminative QA candidates.

\textbf{EMR reasoning-chain construction.}
For EMR diagnosis data, we construct reasoning chains that connect clinical-record evidence to pulmonary diagnoses. Given an EMR case $x$ and reference diagnosis $d$, pulmonary evidence cues are identified from the record and mapped to LungKG entities:
\begin{equation}
A_x = \operatorname{Match}(x, \mathrm{LungKG}),
\end{equation}
where $\operatorname{Match}(\cdot)$ denotes entity and evidence matching between the EMR record and LungKG, and $A_x$ denotes the set of matched pulmonary anchors. We then expand relevant graph neighborhoods around $A_x$ and $d$ to obtain a constrained evidence graph:
\begin{equation}
g_x = \operatorname{Expand}(A_x, d; \mathrm{LungKG}),
\end{equation}
where $\operatorname{Expand}(\cdot)$ retrieves entities, relations, and paths relevant to the matched anchors and reference diagnosis. Conditioned on the EMR case, diagnosis, and serialized evidence graph, the generation procedure constructs a natural-language diagnostic reasoning chain:
\begin{equation}
r_x = \operatorname{Gen}_{\mathrm{EMR\text{-}CoT}}(x, d, \operatorname{Serialize}(g_x)).
\end{equation}

The final training target contains $r_x$ followed by diagnosis $d$. The reasoning chain must use graph-supported entities and relations, preserve relation direction, and avoid unsupported treatment or contraindication claims. This construction is used only for training-data generation; during evaluation, the model receives only the task input and does not access the reference diagnosis.

\paragraph{SFT formatting.}
Both KGQA and EMR samples are converted into an input--target SFT format. For KGQA, \textit{Input} is the generated pulmonary question and \textit{Target} is the LungKG-constrained reasoning chain followed by the final answer. For EMR diagnosis, \textit{Input} is the de-identified clinical record and \textit{Target} is the diagnostic reasoning chain followed by the final diagnosis. LungKG evidence is used during reasoning-chain construction and filtering, rather than exposed as an inference-time retrieval context. This stage teaches Lung-R1 to internalize graph-grounded pulmonary reasoning patterns for record-grounded diagnosis.

\subsection{KG-Guided Reinforcement Learning}

\textbf{Reinforcement learning setup.}
After SFT, Lung-R1 is further optimized with KG-guided reinforcement learning to align outputs with both diagnostic correctness and LungKG-supported reasoning. RL prompts are sampled from non-evaluation KGQA and EMR training sources, with held-out evaluation cases excluded. For each prompt $x$, the sampling policy $\pi_{\theta_{\mathrm{old}}}$ generates candidate outputs $\{y_i\}_{i=1}^{G}$, each scored by a KG-guided reward $R_i=R(y_i,x)$. Group-normalized advantages are computed as:
\begin{equation}
\small
\hat{A}_i =
\frac{R_i-\mu_R}{\sigma_R+\epsilon_{\mathrm{std}}},
\end{equation}
where $\mu_R$ and $\sigma_R$ are the group-level reward mean and standard deviation. Following GRPO~\citep{guo2025deepseek}, the policy is updated with a clipped objective:
\begin{equation}
\small
\mathcal{J}(\theta)=
\mathbb{E}_{x,y_i}
\left[
\min
\left(
r_i \hat{A}_i,
\operatorname{clip}(r_i,1-\epsilon_{\mathrm{clip}},1+\epsilon_{\mathrm{clip}})\hat{A}_i
\right)
\right],
\end{equation}
where $r_i=\pi_\theta(y_i|x)/\pi_{\theta_{\mathrm{old}}}(y_i|x)$ is the importance sampling ratio.

\textbf{KG-guided reward composition.}
The KG-guided reward evaluates both outcome-level diagnostic correctness and process-level consistency with LungKG evidence. It consists of three normalized components.

\textit{1) Diagnosis correctness reward.}
Let $s_{\mathrm{dx}}\in\{0,\ldots,5\}$ denote the diagnosis-alignment score between the final answer and the reference diagnosis under the fixed diagnosis rubric. We normalize it as:
\begin{equation}
\small
R_{\mathrm{dx}}(y_i)=\frac{s_{\mathrm{dx}}}{5}.
\end{equation}
This component anchors the optimization toward clinically correct record-grounded diagnosis.

\textit{2) Graph faithfulness reward.}
For the generated reasoning chain, let $N_{\mathrm{claim}}$ be the number of extracted medical claims, $N_{\mathrm{sup}}$ the number of claims fully supported by the relevant LungKG evidence graph, and $N_{\mathrm{partial}}$ the number of partially supported or synonym-equivalent claims. We define:
\begin{equation}
\small
R_{\mathrm{graph}}(y_i)=
\frac{
N_{\mathrm{sup}} + 0.5N_{\mathrm{partial}}
}{
\max(N_{\mathrm{claim}},1)
}.
\end{equation}
This component encourages the reasoning process to remain grounded in LungKG-supported pulmonary evidence.

\textit{3) Relation/path consistency reward.}
Let $N_{\mathrm{path}}$ be the number of LungKG relation or path steps extracted from the response, and $N_{\mathrm{valid}}$ the number that preserve valid relation types and directed paths, such as disease--symptom, pathogen--infection, examination--diagnosis, drug--contraindication, and treatment--condition relations. We compute:
\begin{equation}
\small
R_{\mathrm{path}}(y_i)=
\frac{
N_{\mathrm{valid}}
}{
\max(N_{\mathrm{path}},1)
}.
\end{equation}
This component rewards relation-consistent reasoning over typed, directed LungKG evidence.

The final reward is a weighted combination of the three components:
\begin{equation}
\small
R_i =
\lambda_{1} R_{\mathrm{dx}}(y_i)
+
\lambda_{2} R_{\mathrm{graph}}(y_i)
+
\lambda_{3} R_{\mathrm{path}}(y_i),
\end{equation}
where $\lambda_{1}$, $\lambda_{2}$, and $\lambda_{3}$ are fixed reward-weight hyperparameters. Their values, reward extraction prompts, graph-matching rules, and scoring details are provided in Appendix~\ref{app:kg_guided_rl_details}. Through this design, LungKG guides RL toward diagnosis correctness, graph-faithful reasoning, and relation-consistent evidence use.

\begin{table*}[t]

\definecolor{TopOne_1}{HTML}{c0bbd8}
\definecolor{TopTwo_1}{HTML}{d7d4e9}
\definecolor{TopThree_1}{HTML}{e7e4f1}
\definecolor{Rest_1}{HTML}{f0eff8}

\definecolor{TopOne_2}{HTML}{ebd0dc}
\definecolor{TopTwo_2}{HTML}{EFE4E8}
\definecolor{TopThree_2}{HTML}{f4edf0}
\definecolor{Rest_2}{HTML}{f8f5fb}

\definecolor{TopOne_3}{HTML}{c8def7}
\definecolor{TopTwo_3}{HTML}{e0e9f7}
\definecolor{TopThree_3}{HTML}{eaf1fa}
\definecolor{Rest_3}{HTML}{f2f7fc}

\definecolor{TopOne_4}{HTML}{cfe7d4}
\definecolor{TopTwo_4}{HTML}{e2f1e4}
\definecolor{TopThree_4}{HTML}{edf7ee}
\definecolor{Rest_4}{HTML}{f6fbf7}

\centering
\tiny
\newcolumntype{S}{>{\raggedright\arraybackslash\tiny}m{3.20cm}}
\newcolumntype{C}{>{\centering\arraybackslash\tiny}m{1.22cm}}
\renewcommand{\arraystretch}{1.3}
\resizebox{\textwidth}{!}{%
\setlength{\tabcolsep}{1.3pt}
\begin{tabular}{S|CCCCCCCCCC}
\hline
\multirow{2}{*}{\textbf{Model}} &
\multicolumn{3}{c}{\textbf{Choice Accuracy}} &
\multicolumn{3}{c}{\textbf{Pulmonary-QA}} &
\multicolumn{4}{c}{\textbf{EMR Diagnosis}} \\
\cline{2-11}
& \textit{Overall (\%)} & \textit{Single (\%)} & \textit{Multi (\%)}
& \textit{Score} & \textit{QWK} & \textit{$\geq$4 Rate (\%)}
& \textit{Score} & \textit{QWK} & \textit{$\geq$4 Rate (\%)} & \textit{Micro R} \\
\hline

\multicolumn{11}{l}{\textit{\textbf{Small Language Models}}} \\
\hline
Qwen2.5-7B
& \cellcolor{Rest_1}58.40
& \cellcolor{Rest_1}70.97
& \cellcolor{Rest_1}21.88
& \cellcolor{Rest_1}3.660
& \cellcolor{Rest_1}0.8863
& \cellcolor{Rest_1}60.40
& \cellcolor{Rest_1}3.7007
& \cellcolor{Rest_1}0.8116
& \cellcolor{Rest_1}47.10
& \cellcolor{Rest_1}0.7281 \\

Qwen2.5-14B
& \cellcolor{Rest_1}63.20
& \cellcolor{TopThree_1}74.19
& \cellcolor{Rest_1}31.25
& \cellcolor{Rest_1}3.996
& \cellcolor{Rest_1}0.8727
& \cellcolor{Rest_1}68.40
& \cellcolor{Rest_1}3.7709
& \cellcolor{Rest_1}0.7322
& \cellcolor{Rest_1}41.50
& \cellcolor{Rest_1}0.7931 \\

Qwen3-30B-A3B
& \cellcolor{Rest_1}64.80
& \cellcolor{TopThree_1}74.19
& \cellcolor{TopTwo_1}37.50
& \cellcolor{TopThree_1}4.292
& \cellcolor{Rest_1}0.8630
& \cellcolor{Rest_1}76.00
& \cellcolor{Rest_1}3.3617
& \cellcolor{Rest_1}0.6904
& \cellcolor{Rest_1}44.70
& \cellcolor{Rest_1}0.8249 \\
\hline

\multicolumn{11}{l}{\textit{\textbf{Large Language Models}}} \\
\hline
Gemini-3-Pro
& \cellcolor{TopTwo_2}67.20
& \cellcolor{TopTwo_2}77.96
& \cellcolor{TopThree_2}35.94
& \cellcolor{TopTwo_2}4.396
& \cellcolor{Rest_2}0.8460
& \cellcolor{TopThree_2}80.80
& \cellcolor{Rest_2}4.0688
& \cellcolor{Rest_2}0.5431
& \cellcolor{Rest_2}50.70
& \cellcolor{Rest_2}0.7056 \\

DeepSeek-V3.2
& \cellcolor{TopThree_2}66.80
& \cellcolor{TopTwo_2}77.96
& \cellcolor{Rest_2}34.38
& \cellcolor{Rest_2}4.256
& \cellcolor{Rest_2}0.8394
& \cellcolor{Rest_2}76.40
& \cellcolor{Rest_2}4.0573
& \cellcolor{Rest_2}0.6371
& \cellcolor{Rest_2}52.40
& \cellcolor{Rest_2}0.8489 \\

DeepSeek-R1
& \cellcolor{Rest_2}63.60
& \cellcolor{Rest_2}71.51
& \cellcolor{TopOne_2}40.62
& \cellcolor{Rest_2}4.264
& \cellcolor{Rest_2}0.8847
& \cellcolor{Rest_2}77.60
& \cellcolor{Rest_2}3.9891
& \cellcolor{Rest_2}0.7129
& \cellcolor{Rest_2}48.90
& \cellcolor{Rest_2}0.8103 \\

Claude-Sonnet-4.5
& \cellcolor{Rest_2}60.40
& \cellcolor{Rest_2}67.20
& \cellcolor{TopOne_2}40.62
& \cellcolor{TopTwo_2}4.396
& \cellcolor{Rest_2}0.9046
& \cellcolor{TopOne_2}82.40
& \cellcolor{TopThree_2}4.2107
& \cellcolor{Rest_2}0.5887
& \cellcolor{TopThree_2}61.90
& \cellcolor{TopThree_2}0.8753 \\

GPT-5-chat
& \cellcolor{Rest_2}52.00
& \cellcolor{Rest_2}61.29
& \cellcolor{Rest_2}25.00
& \cellcolor{Rest_2}4.1551
& \cellcolor{Rest_2}0.9002
& \cellcolor{Rest_2}75.10
& \cellcolor{Rest_2}4.1986
& \cellcolor{Rest_2}0.5990
& \cellcolor{Rest_2}56.00
& \cellcolor{Rest_2}0.8593 \\

GPT-5.1
& \cellcolor{Rest_2}54.00
& \cellcolor{Rest_2}61.29
& \cellcolor{Rest_2}32.81
& \cellcolor{Rest_2}4.1488
& \cellcolor{TopTwo_2}0.9182
& \cellcolor{Rest_2}73.14
& \cellcolor{Rest_2}4.2075
& \cellcolor{Rest_2}0.6120
& \cellcolor{Rest_2}60.50
& \cellcolor{Rest_2}0.7868 \\

GPT-4o
& \cellcolor{Rest_2}49.60
& \cellcolor{Rest_2}58.06
& \cellcolor{Rest_2}25.00
& \cellcolor{Rest_2}4.176
& \cellcolor{Rest_2}0.8611
& \cellcolor{Rest_2}72.80
& \cellcolor{Rest_2}3.9908
& \cellcolor{Rest_2}0.6758
& \cellcolor{Rest_2}53.50
& \cellcolor{Rest_2}0.8493 \\

GPT-4.1
& \cellcolor{Rest_2}50.40
& \cellcolor{Rest_2}59.68
& \cellcolor{Rest_2}23.44
& \cellcolor{Rest_2}4.132
& \cellcolor{Rest_2}0.8739
& \cellcolor{Rest_2}72.80
& \cellcolor{Rest_2}4.0684
& \cellcolor{Rest_2}0.7178
& \cellcolor{Rest_2}57.30
& \cellcolor{Rest_2}0.8606 \\

Kimi-K2-Thinking
& \cellcolor{Rest_2}44.40
& \cellcolor{Rest_2}55.38
& \cellcolor{Rest_2}12.50
& \cellcolor{Rest_2}3.440
& \cellcolor{Rest_2}0.8749
& \cellcolor{Rest_2}52.80
& \cellcolor{Rest_2}3.3112
& \cellcolor{Rest_2}0.8266
& \cellcolor{Rest_2}31.30
& \cellcolor{Rest_2}0.6844 \\
\hline

\multicolumn{11}{l}{\textit{\textbf{Medical-specific Large Language Models}}} \\
\hline
ClinicalGPT-R1
& \cellcolor{Rest_3}59.20
& \cellcolor{Rest_3}73.12
& \cellcolor{Rest_3}18.75
& \cellcolor{Rest_3}3.588
& \cellcolor{Rest_3}0.8598
& \cellcolor{Rest_3}55.20
& \cellcolor{Rest_3}3.7449
& \cellcolor{Rest_3}0.7452
& \cellcolor{Rest_3}47.60
& \cellcolor{Rest_3}0.7958 \\

HuatuoGPT-o1-7B
& \cellcolor{Rest_3}62.80
& \cellcolor{TopTwo_3}77.96
& \cellcolor{Rest_3}18.75
& \cellcolor{Rest_3}3.636
& \cellcolor{Rest_3}0.8686
& \cellcolor{Rest_3}58.40
& \cellcolor{Rest_3}3.3687
& \cellcolor{TopTwo_3}0.8278
& \cellcolor{Rest_3}37.40
& \cellcolor{Rest_3}0.6578 \\

MedGemma-27B
& \cellcolor{Rest_3}45.20
& \cellcolor{Rest_3}55.38
& \cellcolor{Rest_3}15.62
& \cellcolor{Rest_3}3.792
& \cellcolor{Rest_3}0.8477
& \cellcolor{Rest_3}65.20
& \cellcolor{Rest_3}3.7347
& \cellcolor{Rest_3}0.6556
& \cellcolor{Rest_3}37.20
& \cellcolor{Rest_3}0.8343 \\

Baichuan-M2
& \cellcolor{Rest_3}62.40
& \cellcolor{Rest_3}71.51
& \cellcolor{TopThree_3}35.94
& \cellcolor{Rest_3}4.016
& \cellcolor{Rest_3}0.8822
& \cellcolor{Rest_3}68.00
& \cellcolor{Rest_3}3.9339
& \cellcolor{Rest_3}0.7271
& \cellcolor{Rest_3}50.80
& \cellcolor{Rest_3}0.8196 \\

GPT-OSS-20B
& \cellcolor{Rest_3}47.60
& \cellcolor{Rest_3}59.14
& \cellcolor{Rest_3}14.06
& \cellcolor{Rest_3}3.768
& \cellcolor{Rest_3}0.8607
& \cellcolor{Rest_3}65.60
& \cellcolor{Rest_3}3.2727
& \cellcolor{Rest_3}0.7026
& \cellcolor{Rest_3}24.20
& \cellcolor{Rest_3}0.7905 \\

GPT-OSS-120B
& \cellcolor{Rest_3}52.00
& \cellcolor{Rest_3}62.37
& \cellcolor{Rest_3}21.88
& \cellcolor{Rest_3}4.108
& \cellcolor{Rest_3}0.8886
& \cellcolor{Rest_3}73.20
& \cellcolor{Rest_3}3.9218
& \cellcolor{Rest_3}0.5735
& \cellcolor{Rest_3}40.80
& \cellcolor{TopTwo_3}0.8795 \\
\hline

\multicolumn{11}{l}{\textit{\textbf{Our Pulmonary LLMs}}} \\
\hline
Lung-R1-7B
& \cellcolor{TopTwo_4}67.20
& \cellcolor{TopTwo_4}77.96
& \cellcolor{Rest_4}35.94
& \cellcolor{Rest_4}4.120
& \cellcolor{TopThree_4}0.9123
& \cellcolor{Rest_4}72.80
& \cellcolor{TopTwo_4}4.3550
& \cellcolor{TopOne_4}0.8295
& \cellcolor{TopOne_4}75.00
& \cellcolor{TopOne_4}0.8939 \\

Lung-R1-14B
& \cellcolor{TopOne_4}\textbf{67.60}
& \cellcolor{TopOne_4}81.72
& \cellcolor{TopThree_4}26.56
& \cellcolor{TopOne_4}\textbf{4.416}
& \cellcolor{TopOne_4}0.9239
& \cellcolor{TopTwo_4}81.20
& \cellcolor{TopOne_4}\textbf{4.3583}
& \cellcolor{TopThree_4}0.8275
& \cellcolor{TopTwo_4}71.30
& \cellcolor{TopOne_4}\textbf{0.8939} \\
\hline

\end{tabular}
}
\caption{Main results of pulmonary evaluation tasks. Colors distinguish model groups; darker cells rank higher. Bold marks the best value among all systems for primary metrics and EMR Micro R.}
\label{tab:main_results}
\end{table*}

\section{Experiments}
\label{sec:experiments}

\subsection{Evaluation Setup}
Given the lack of a unified pulmonary evaluation suite covering examination-style knowledge, physician-reviewed QA, and record-grounded EMR diagnosis, we construct a held-out suite with three tasks: Choice, Pulmonary-QA, and EMR Diagnosis. It contains 250 Choice questions, 250 physician-reviewed Pulmonary-QA items, and 300 de-identified EMR Diagnosis cases. Choice and Pulmonary-QA are built from pulmonary examination-style and physician-reviewed knowledge sources\footnote{\url{https://www.medtiku.com/}.}, while EMR Diagnosis cases come from de-identified clinical records strictly non-overlapping with EMR training samples. Pulmonary clinicians reviewed the annotations and scoring rubrics; EMR Diagnosis uses a 0--5 diagnosis-alignment rubric. We treat EMR Diagnosis as the primary task because it directly tests whether pulmonary knowledge transfers to case-level diagnostic reasoning. Detailed construction, statistics, annotation procedures, and rubrics are provided in Appendix~\ref{app:evaluation_details} and Appendix~\ref{app:evaluation_task_characterization}.

We compare 20 systems across 4 groups (Figure~\ref{fig:lungr1_pipeline}(c)): Small language models include Qwen2.5-7B/14B~\citep{qwen2025qwen25} and Qwen3-30B-A3B~\citep{qwen2025qwen3}. General-purpose LLMs include Gemini-3-Pro~\citep{googledeepmind2025gemini3pro}, DeepSeek-V3.2~\citep{deepseekai2025v32exp}, DeepSeek-R1~\citep{deepseekai2025r1}, Claude-Sonnet-4.5~\citep{anthropic2025claudesonnet45}, GPT-5-chat~\citep{openai2025gpt5system}, GPT-5.1~\citep{openai2025gpt51}, GPT-4o~\citep{openai2024gpt4o}, GPT-4.1~\citep{openai2025gpt41}, and Kimi-K2-Thinking~\citep{moonshot2025kimik2thinking}. Medical-specific LLMs include ClinicalGPT-R1~\citep{lan2025clinicalgptr1}, HuatuoGPT-o1-7B~\citep{chen2024huatuogpto1}, MedGemma-27B~\citep{sellergren2025medgemma}, Baichuan-M2~\citep{baichuanm2team2025baichuanm2}, and GPT-OSS-20B/120B~\citep{agarwal2025gptoss}. The remaining two systems are Lung-R1 variants.

All models follow identical prompting, decoding and scoring rules. We use overall and subtype accuracy for the Choice task~\citep{powers2020evaluation}. For Pulmonary-QA and EMR Diagnosis, we report mean 0--5 rubric score, score-\(\geq4\) rate~\citep{zheng2023judging} and QWK for consistency with physician ratings~\citep{cohen1968weighted}; EMR Diagnosis additionally adopts Micro R for deterministic diagnosis matching~\citep{sokolova2009systematic}. For reproducibility, Lung-R1 is built on Qwen2.5-7B/14B-Instruct, with a learning rate of \(2\times10^{-6}\) and global batch size 64. Full model identifiers, metric definitions, evaluation and training configurations are provided in Appendix~\ref{app:evaluation_details} and Appendix~\ref{app:training_config}.

\subsection{Main Results}

Table~\ref{tab:main_results} reports the main comparison across the three evaluation tasks. The primary task-level metrics are overall Choice accuracy, Pulmonary-QA mean score, and EMR Diagnosis score. Auxiliary metrics include QWK, score-$\geq4$ rate, Choice subtype accuracy, and EMR Micro R; these measure score agreement, high-score output concentration, and deterministic diagnosis matching rather than defining a single aggregate score.

In the 20-system comparison, Lung-R1-14B obtains the strongest performance on all three primary task-level metrics. Unless otherwise specified, the Lung-R1 rows in the main comparison report the final checkpoints after KG-guided RL. These results suggest that LungKG-guided training is effective not only for pulmonary knowledge checks but also for record-grounded diagnosis, the task most directly tied to the Pulmonary Knowledge-to-Diagnosis Gap.

\subsection{EMR Diagnosis and Clinician Validation}

We further analyze EMR Diagnosis because it is the primary record-grounded evaluation. On the 300 held-out de-identified EMR cases, each model receives only the clinical record and must generate the pulmonary diagnosis without access to the reference diagnosis. Outputs are scored using the 0--5 diagnosis-alignment rubric, and Figure~\ref{fig:emr_score_distribution} shows the score-bin distribution for representative systems.

For EMR Diagnosis, Lung-R1 variants achieve the top two mean scores and concentrate more cases in the highest-score bin. Some baselines remain competitive on auxiliary metrics, so we interpret Lung-R1's results as strong task-level performance rather than dominance on every secondary metric. A detailed Lung-R1 discrepancy analysis is provided in Appendix~\ref{app:error_analysis}.

For clinician validation, pulmonary clinicians review de-identified records, reference diagnoses, and anonymized model outputs using the same 0--5 rubric, with disagreements resolved by adjudication. The physician-adjudicated ratings serve as the human reference for QWK computation. As shown in Table~\ref{tab:main_results}, Lung-R1 variants show high agreement with clinician ratings, with QWK exceeding 0.90 on Pulmonary-QA and 0.82 on EMR Diagnosis. We therefore use QWK as score-agreement evidence rather than direct proof of clinical correctness; detailed instructions are provided in Appendix~\ref{app:evaluation_details}.

\begin{figure}[t]
  \centering
  \includegraphics[width=\columnwidth]{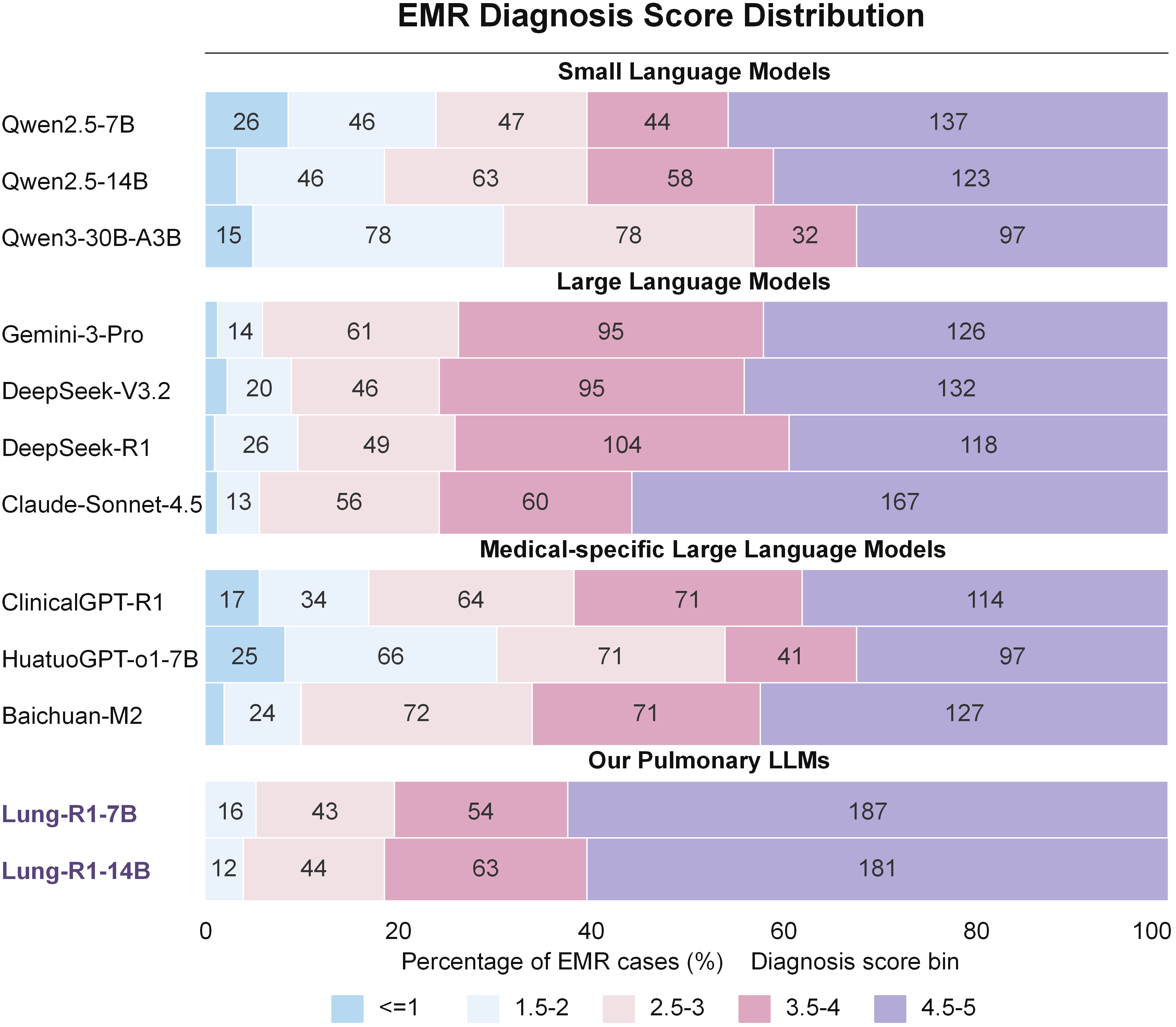}
  \caption{Distribution of EMR Diagnosis scores. Lung-R1 variants show a higher concentration of cases in the highest-score bin.}
  \label{fig:emr_score_distribution}
\end{figure}

\subsection{Ablation Studies}

We analyze three factors: SFT supervision source, explicit CoT format, and KG-guided reinforcement learning. Table~\ref{tab:ablation_all} summarizes the results. In the source block, \textit{w/o EMR} and \textit{w/o KGQA} denote KGQA-only and EMR-only SFT, respectively, relative to the full mixed SFT setting. In the CoT block, \textit{w/o CoT} removes explicit reasoning chains while keeping the same task inputs and final-answer targets. \textit{KG-RL} denotes the KG-guided reinforcement-learning stage after SFT.

\begin{table}[t]
\centering
\footnotesize
\setlength{\tabcolsep}{3pt}
\renewcommand{\arraystretch}{1.06}

\begin{tabularx}{\columnwidth}{@{}>{\raggedright\arraybackslash}Xccc@{}}
\toprule
\textbf{Variant} & \textbf{Choice} & \textbf{P-QA} & \makecell{\textbf{EMR}\\\textbf{Dx}} \\
\midrule
\multicolumn{4}{@{}l}{\textit{\textbf{SFT source ablation, 7B}}} \\
Base-7B
& 58.4 & 3.660 & 3.7007 \\
7B w/o EMR
& 61.2 & 4.084 & 3.4949 \\
7B w/o KGQA
& 62.0 & 3.664 & 4.2626 \\
7B Full SFT
& 64.1 & 4.112 & 4.3471 \\
\midrule

\multicolumn{4}{@{}l}{\textit{\textbf{CoT-format ablation, SFT only}}} \\
7B w/o CoT
& 63.8 & 4.172 & 4.1918 \\
7B w/ CoT
& 64.1  & 4.112 & 4.3471 \\
14B w/o CoT
& 65.5 & 4.244 & 4.3041 \\
14B w/ CoT
& 67.2 & 4.298 & 4.3575 \\
\midrule

\multicolumn{4}{@{}l}{\textit{\textbf{KG-guided RL ablation, 14B CoT}}} \\
14B w/o KG-RL
& 67.2 & 4.298 & 4.3575 \\
14B w/ KG-RL
& 67.6 & 4.416 & 4.3583 \\
\bottomrule
\end{tabularx}

\caption{Ablation study on SFT source, CoT format, and KG-guided RL.}
\label{tab:ablation_all}
\end{table}

The source block shows that KGQA and EMR supervision are complementary: KGQA improves Pulmonary-QA, EMR provides stronger diagnosis grounding, and Full SFT achieves the best 7B results across all three metrics. The CoT block shows that explicit reasoning supervision improves Choice and EMR Diagnosis for both backbone sizes, although Pulmonary-QA gains are not uniform. The RL block shows that KG-guided RL improves Choice and Pulmonary-QA while preserving strong EMR Diagnosis performance, suggesting that reward alignment strengthens pulmonary knowledge performance without degrading record-grounded diagnosis.

\subsection{Case Study}
Figure~\ref{fig:emr_case_misalignment} provides a qualitative illustration rather than a systematic error taxonomy. In this de-identified case, Lung-R1 better follows the acute infectious respiratory-failure pattern, while Claude-Sonnet-4.5 emphasizes secondary imaging or cardiopulmonary findings. The example illustrates the need to prioritize clinically central evidence in record-grounded pulmonary diagnosis. Additional qualitative examples are provided in Appendix~\ref{app:case_study}.

\begin{figure}[t]
  \centering
  \includegraphics[width=\columnwidth]{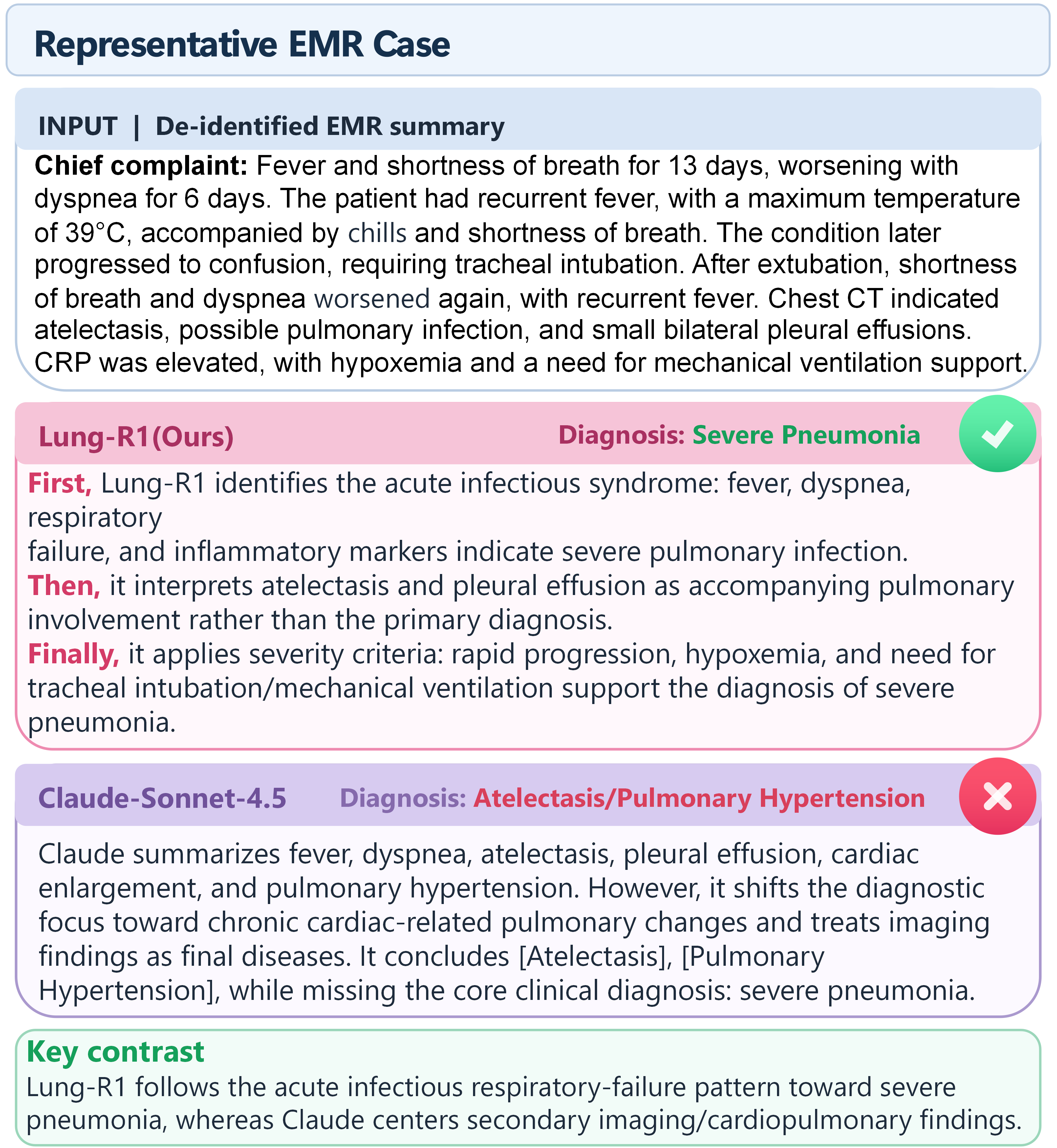}
  \caption{Qualitative illustration of one de-identified EMR Diagnosis case comparing Claude-Sonnet-4.5 and Lung-R1-14B. Lung-R1 better aligns with the reference diagnosis of severe pneumonia.}
  \label{fig:emr_case_misalignment}
\end{figure}

\section{Conclusion}

This work addresses the \textit{Pulmonary Knowledge-to-Diagnosis Gap} between pulmonary knowledge acquisition and record-grounded clinical diagnosis. We introduce \textbf{LungKG}, a structured pulmonary knowledge graph that organizes fragmented pulmonary knowledge into typed, directed relations for model training and alignment. Built on LungKG, \textbf{Lung-R1} learns record-grounded pulmonary diagnostic reasoning through KG-constrained Chain-of-Thought supervision and KG-guided reinforcement learning. In our 20-system evaluation, Lung-R1 achieves state-of-the-art primary-metric performance, with the clearest gains on EMR Diagnosis. These findings show that LungKG-guided training offers an effective path toward clinically grounded, specialty-specific medical LLMs, advancing reliable AI-assisted diagnosis from knowledge recall to record-grounded clinical reasoning.

\section*{Limitations}
Despite the promising results, this study has limitations inherent to diagnosis-oriented pulmonary reasoning. Our current evaluation setting is primarily text-based, whereas real-world pulmonary diagnosis may involve multimodal and longitudinal evidence, including raw imaging, temporal disease trajectories, bedside findings, and treatment response. Although Lung-R1 improves record-grounded pulmonary diagnosis, it is not a complete clinical decision-making system; the current tasks focus on diagnosis rather than treatment planning, medication selection, safety judgment, or patient management, and all clinical outputs require qualified physician review. The evaluation suite is not designed as a class-balanced disease taxonomy: Choice, Pulmonary-QA, and EMR Diagnosis reflect the source distributions and clinical settings used in this study, and the EMR ICD-10 statistics reported in the appendix are counted over diagnosis mentions rather than cases. 
Future work could incorporate multimodal and treatment-oriented evaluation and combine automatic scoring with broader physician assessment.

\section*{Ethics Statement}
\label{ethics_statement}
This study was conducted in accordance with ethical guidelines for clinical research and the principles of the \textit{Declaration of Helsinki}. The study protocol, including data collection, de-identification, annotation, LungKG construction, model training, and evaluation procedures, was reviewed and approved through an institutional ethics review process. This review process covered the use of retrospective pulmonary electronic medical records, clinical annotation procedures, evaluation-task construction, LungKG-guided training, and downstream model assessment.

Data collection and annotation activities were designed to protect patient privacy and maintain the integrity of the research process. All pulmonary electronic medical records used in this study were de-identified prior to any processing, annotation, model training, or evaluation. Patient-identifying information was removed before research use, and no institution names, ethics approval identifiers, physician identities, local data paths, or other potentially identifying governance details are disclosed in the anonymous-review version. Because the study used de-identified retrospective records and posed minimal privacy risk, the requirement for individual informed consent was waived under the approved ethics review process.

Clinical annotation, LungKG-guided reasoning-chain construction, and model evaluation were conducted solely for research purposes. The annotation process followed standardized pulmonary review criteria, and all clinical labels, graph-derived supervision, and review outputs were used only for evaluation-task construction, model training, and research analysis. The study does not release identifiable patient information.

Lung-R1 is a research model for pulmonary language-model evaluation and diagnosis-oriented adaptation. It is not intended for direct clinical decision-making without further clinical validation and regulatory approval. Any diagnostic or treatment-related output generated by the model must be reviewed by qualified clinicians, and human oversight remains necessary in high-stakes medical settings.

For transparency, we note that the data resources and study procedures were reviewed and approved by the relevant institutional ethics review process under formal protocols. To preserve double-blind anonymity, detailed ethics approval identifiers and data-governance information are omitted from the anonymous-review version. Where permitted, relevant institutional and governance details can be provided in the final version or under controlled data-access procedures.

\bibliography{references}

\appendix

\section{Additional Benchmark and Method Details}

\subsection{Evaluation Details}
\label{app:evaluation_details}

\paragraph{Baselines.}
We evaluate 20 systems spanning four categories. Table~\ref{tab:main_results} uses reader-facing model names; exact evaluated API/model identifiers are shown in \texttt{monospace} where they differ. Small language models include Qwen2.5-7B and Qwen2.5-14B~\citep{qwen2025qwen25}, and Qwen3-30B-A3B~\citep{qwen2025qwen3}. General large language models include Gemini-3-Pro~\citep{googledeepmind2025gemini3pro}, DeepSeek-V3.2~\citep{deepseekai2025v32exp}, DeepSeek-R1~\citep{deepseekai2025r1}, Claude-Sonnet-4.5~\citep{anthropic2025claudesonnet45}, GPT-5-chat and GPT-5.1~\citep{openai2025gpt5system,openai2025gpt51}, GPT-4o~\citep{openai2024gpt4o}, GPT-4.1~\citep{openai2025gpt41}, and Kimi-K2-Thinking~\citep{moonshot2025kimik2thinking}. Medical-specific large language models include ClinicalGPT-R1~\citep{lan2025clinicalgptr1}, HuatuoGPT-o1-7B~\citep{chen2024huatuogpto1}, MedGemma-27B~\citep{sellergren2025medgemma}, Baichuan-M2~\citep{baichuanm2team2025baichuanm2}, and GPT-OSS-20B/120B~\citep{agarwal2025gptoss}. Our pulmonary LLMs are Lung-R1-7B/14B.

\paragraph{Metrics and protocol.}
For Choice, we report overall accuracy over the 250-question set, with single-answer and multi-answer accuracies as auxiliary breakdowns. For Pulmonary-QA and EMR Diagnosis, we used five frozen LLM-as-Judge models: DeepSeek-V4-Pro~\citep{deepseekai2026deepseekv4}, Kimi-K2.6~\citep{moonshot2026kimik26}, GPT-5.5~\citep{openai2026gpt55}, Claude-Opus-4.6~\citep{anthropic2026claudeopus46}, and Gemini-3.1-Pro~\citep{googledeepmind2026gemini31pro}. Each judge scored every QA and EMR model output under the same task-specific rubric, fixed prompts, and a fixed decoding configuration with temperature set to 0.1; no model-specific prompt tuning was applied. For Pulmonary-QA, model outputs are compared against reference answers and scored on a 0--5 scale. For EMR Diagnosis, model outputs are scored against reference diagnoses and also evaluated with micro-averaged deterministic diagnosis recall.

For each scored QA or EMR output $n$, let $s_{n,j} \in \{0,\ldots,5\}$ denote the score assigned by judge $j \in \{1,\ldots,5\}$. The ensemble score is $\bar{s}_n=\frac{1}{5}\sum_{j=1}^{5}s_{n,j}$. We report the mean QA or EMR score by averaging $\bar{s}_n$ over samples. The score-$\geq4$ rate is the proportion of samples with $\bar{s}_n \geq 4$. We average the raw 0--5 judge scores before any discretization; discretization is used only when converting $\bar{s}_n$ to the ordinal automatic score for QWK. Any unparseable or out-of-range judge response is flagged and re-scored under the same fixed prompt and decoding configuration before aggregation, so only valid 0--5 scores enter the ensemble.

For QWK, each evaluated model output is paired at the sample level with two ordinal scores for the same answer: a physician-adjudicated human score $h_n \in \{0,\ldots,5\}$ and an ensemble automatic score $a_n=\min(5,\max(0,\mathrm{round}(\bar{s}_n)))$. Let $O_{ij}$ denote the observed count of pairs with physician score $i$ and ensemble automatic score $j$, and let $E_{ij}$ denote the expected count from the marginal score distributions. With $K=6$ score levels and quadratic weights $w_{ij}=(i-j)^2/(K-1)^2$, we compute
\begin{equation}
\kappa_w =
1 -
\frac{\sum_{i=0}^{5}\sum_{j=0}^{5} w_{ij}O_{ij}}
{\sum_{i=0}^{5}\sum_{j=0}^{5} w_{ij}E_{ij}} .
\label{eq:qwk}
\end{equation}
Higher QWK indicates stronger agreement between ensemble automatic scoring and physician-adjudicated ratings, not proof of clinical correctness.

Table~\ref{tab:main_results} reports the expanded 20-system comparison, including Choice sub-type accuracies, Pulmonary-QA score and agreement diagnostics, and EMR Diagnosis score plus deterministic recall. We keep these diagnostics in the main table to avoid duplicating result tables in the appendix.

\subsection{Evaluation Task Characterization}
\label{app:evaluation_task_characterization}

This appendix reports task-level characterization statistics for the pulmonary evaluation suite. Figure~\ref{fig:evaluation_task_characterization} summarizes the task composition, while the following tables provide the complete source, topic, and diagnosis-group distributions. Choice provenance is computed from the frozen Choice metadata. Pulmonary-QA topic labels are LLM-assisted characterization labels produced with DeepSeek-V4-Pro; they are used to describe topic coverage and should not be read as physician topic annotations. EMR ICD-10 groups are counted over diagnosis mentions rather than cases because one case can contain more than one reference diagnosis.

\begin{figure}[h]
  \centering
  \includegraphics[width=\columnwidth]{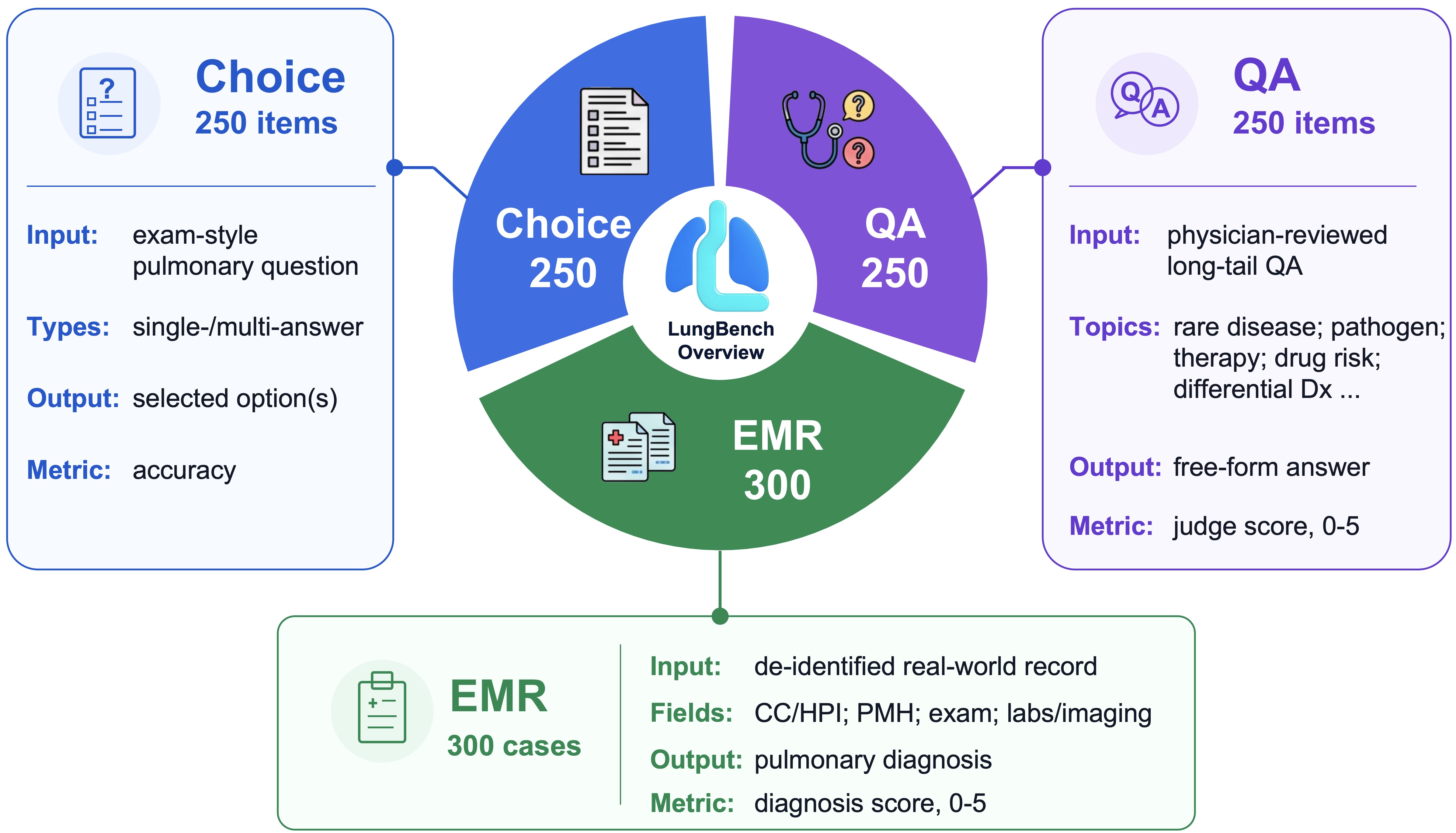}
  \caption{Overview of the pulmonary evaluation tasks, including Choice, Pulmonary-QA, and EMR Diagnosis components.}
  \label{fig:evaluation_task_characterization}
\end{figure}

\begin{table}[h]
  \centering
  \small
  \setlength{\tabcolsep}{4pt}
  \renewcommand{\arraystretch}{1.05}
  \caption{Choice subset characterization.}
  \resizebox{\columnwidth}{!}{%
    \begin{tabular}{lrr}
    \toprule
    \textbf{Choice provenance or format} & \textbf{\#Items} & \textbf{Percent} \\
    \midrule
    Physician examination & 140 & 56.0\% \\
    Professional knowledge & 46 & 18.4\% \\
    Nursing examination & 26 & 10.4\% \\
    Medical technology examination & 19 & 7.6\% \\
    Medical postgraduate examination & 10 & 4.0\% \\
    Pharmacist examination & 9 & 3.6\% \\
    \midrule
    Single-answer MCQ & 186 & 74.4\% \\
    Multi-answer MCQ & 64 & 25.6\% \\
    \bottomrule
    \end{tabular}%
  }
  \label{tab:evaluation_choice_characterization}
\end{table}

\begin{table}[h]
  \centering
  \small
  \setlength{\tabcolsep}{4pt}
  \renewcommand{\arraystretch}{1.05}
  \caption{Most frequent Choice exam classes.}
  \resizebox{\columnwidth}{!}{%
    \begin{tabular}{lrr}
    \toprule
    \textbf{Exam class} & \textbf{\#Items} & \textbf{Percent} \\
    \midrule
    Intermediate professional title & 66 & 26.4\% \\
    Licensed physician & 27 & 10.8\% \\
    Clinical medicine & 26 & 10.4\% \\
    Basic medicine & 19 & 7.6\% \\
    Senior professional title & 18 & 7.2\% \\
    Assistant licensed physician & 16 & 6.4\% \\
    Western medicine comprehensive exam & 13 & 5.2\% \\
    Standardized residency completion & 12 & 4.8\% \\
    Supervising nurse & 10 & 4.0\% \\
    Supervising technician & 10 & 4.0\% \\
    \bottomrule
    \end{tabular}%
  }
  \label{tab:evaluation_choice_classes}
\end{table}

\begin{table}[h]
  \centering
  \small
  \setlength{\tabcolsep}{4pt}
  \renewcommand{\arraystretch}{1.05}
  \caption{LLM-assisted Pulmonary-QA topic characterization using \texttt{DeepSeek-V4-Pro}.}
  \resizebox{\columnwidth}{!}{%
    \begin{tabular}{lrr}
    \toprule
    \textbf{Topic} & \textbf{\#Items} & \textbf{Percent} \\
    \midrule
    Drug / treatment risk & 151 & 60.4\% \\
    Diagnosis / differential diagnosis & 40 & 16.0\% \\
    Pathogen / infection & 35 & 14.0\% \\
    Exam / imaging & 21 & 8.4\% \\
    Rare / specialty disease & 2 & 0.8\% \\
    Other & 1 & 0.4\% \\
    \bottomrule
    \end{tabular}%
  }
  \label{tab:evaluation_pqa_topics}
\end{table}

\begin{table}[h]
  \centering
  \small
  \setlength{\tabcolsep}{4pt}
  \renewcommand{\arraystretch}{1.05}
  \caption{EMR Diagnosis ICD-10 reference-diagnosis groups counted over diagnosis mentions.}
  \resizebox{\columnwidth}{!}{%
    \begin{tabular}{lrr}
    \toprule
    \textbf{ICD-10 group} & \textbf{\#Mentions} & \textbf{Percent} \\
    \midrule
    J09--J18 Influenza and pneumonia & 244 & 64.7\% \\
    J40--J47 Chronic lower respiratory diseases & 72 & 19.1\% \\
    Non-J / outside ICD-10 Chapter X & 35 & 9.3\% \\
    J95--J99 Other respiratory diseases & 15 & 4.0\% \\
    J90--J94 Pleural diseases & 5 & 1.3\% \\
    J60--J70 Lung diseases due to external agents & 3 & 0.8\% \\
    J85--J86 Suppurative and necrotic conditions & 2 & 0.5\% \\
    J20--J22 Other acute lower respiratory infections & 1 & 0.3\% \\
    \bottomrule
    \end{tabular}%
  }
  \label{tab:evaluation_emr_icd10}
\end{table}

\begin{table}[h]
  \centering
  \small
  \setlength{\tabcolsep}{4pt}
  \renewcommand{\arraystretch}{1.05}
  \caption{Additional EMR Diagnosis characterization statistics.}
  \resizebox{\columnwidth}{!}{%
    \begin{tabular}{lrr}
    \toprule
    \textbf{Statistic} & \textbf{Value} & \textbf{Unit} \\
    \midrule
    Cases with one reference diagnosis & 236 & cases \\
    Cases with two reference diagnoses & 51 & cases \\
    Cases with three reference diagnoses & 13 & cases \\
    Mean reference diagnoses per case & 1.26 & diagnoses \\
    Minimum record length & 331 & characters \\
    Median record length & 751.5 & characters \\
    Maximum record length & 2,968 & characters \\
    \bottomrule
    \end{tabular}%
  }
  \label{tab:evaluation_emr_aux}
\end{table}

\begin{table}[h]
  \centering
  \small
  \setlength{\tabcolsep}{4pt}
  \renewcommand{\arraystretch}{1.05}
  \caption{Most frequent normalized EMR Diagnosis reference-diagnosis strings.}
  \resizebox{\columnwidth}{!}{%
    \begin{tabular}{lrr}
    \toprule
    \textbf{Reference diagnosis} & \textbf{\#Mentions} & \textbf{Percent} \\
    \midrule
    Pulmonary infection & 136 & 36.1\% \\
    Acute exacerbation of chronic obstructive pulmonary disease & 56 & 14.9\% \\
    Severe pneumonia & 28 & 7.4\% \\
    Pneumonia & 28 & 7.4\% \\
    Community-acquired pneumonia & 19 & 5.0\% \\
    Bilateral pneumonia & 15 & 4.0\% \\
    Pulmonary shadow & 11 & 2.9\% \\
    Type-II respiratory failure & 8 & 2.1\% \\
    Viral pneumonia & 8 & 2.1\% \\
    Bronchopneumonia & 6 & 1.6\% \\
    \bottomrule
    \end{tabular}%
  }
  \label{tab:evaluation_emr_top_diagnoses}
\end{table}

\subsection{Clinical Annotation Team and Review Protocol}
\label{app:annotation_team}

Evaluation-task annotation was conducted by an anonymized clinical team consisting of 10 licensed pulmonary clinicians and one senior pulmonary clinician adjudicator. The 10-member clinician panel reviewed pulmonary QA items and EMR diagnosis labels against standardized pulmonary review criteria and task-specific rubrics. Disagreements or uncertain cases were escalated to the senior adjudicator for final resolution.

For Pulmonary-QA, the annotation process checked answer correctness, specialty relevance, and whether each item reflected clinically meaningful pulmonary knowledge. For EMR Diagnosis, clinicians reviewed the de-identified record fields and reference pulmonary diagnoses. For QWK agreement analysis, physicians additionally scored the same model outputs under the 0--5 rubrics, and the adjudicated human scores provide the physician-side ordinal scores used for comparison with ensemble automatic scores. We report this team composition only in anonymized form for double-blind review and do not disclose names, institutions, departments, license identifiers, approval identifiers, or local data paths.

\begin{table}[!htbp]
  \centering
  \renewcommand{\arraystretch}{1.1}
  \caption{Key training hyperparameters for Lung-R1.}
  \resizebox{\columnwidth}{!}{%
    \begin{tabular}{lcc}
    \toprule
    \textbf{Parameter} & \textbf{7B} & \textbf{14B} \\
    \midrule
    Method & Instruction tuning & Instruction tuning \\
    Learning rate & $2\times10^{-6}$ & $2\times10^{-6}$ \\
    Epochs & 3 & 2 \\
    Cutoff length & 2048 & 3072 \\
    Per-device batch size & 4 & 2 \\
    Gradient accumulation & 8 & 8 \\
    Global batch size & 64 & 64 \\
    Optimizer & AdamW & AdamW \\
    LR schedule / warmup & Cosine / 0.03 & Cosine / 0.03 \\
    Precision & BF16 & BF16 \\
    Distributed training & ZeRO-3 & ZeRO-3 \\
    Save / eval steps & 200 / 200 & 100 / 100 \\
    Hardware & 2 $\times$ H800 80G & 4 $\times$ H800 80G \\
    \bottomrule
    \end{tabular}%
  }
  \label{tab:training_config}
\end{table}

\subsection{EMR Diagnosis Error and Discrepancy Analysis}
\label{app:error_analysis}

To better understand the failure modes of EMR Diagnosis outputs, we manually analyze mismatched or unverified predictions and categorize them into five discrepancy types. Table~\ref{tab:error_types} summarizes the taxonomy and distribution.

\textbf{Semantic Equivalence (SE)} denotes cases where the model uses a clinically equivalent synonym rather than the exact reference term, such as ``pulmonary infection'' instead of ``pneumonia.'' 
\textbf{Overprediction (OP)} refers to outputs that list multiple possible diagnoses without clearly identifying the primary diagnosis, or that over-generate secondary conditions and complications.
\textbf{Missed Diagnosis (MD)} indicates failure to identify the core diagnosis in the reference answer.
\textbf{Splitting / Combining Diagnoses (SD)} refers to cases where the model improperly splits a composite diagnosis or combines distinct clinical conditions.
\textbf{Wrong Inference (WI)} denotes cases where the model recognizes partial clinical evidence but draws an incorrect primary diagnostic conclusion.

Among 232 categorized discrepancies, the most frequent type is SE, accounting for 83 cases (35.78\%), followed by OP with 79 cases (34.05\%). Together, SE and OP account for 69.83\% of discrepancies, suggesting that many raw mismatches arise from terminology variation or over-enumerated diagnostic outputs rather than complete diagnostic failure. In contrast, MD accounts for 58 cases (25.00\%), representing the main clinically meaningful failure mode: the model fails to isolate the primary disease from complex clinical narratives. SD accounts for 10 cases (4.31\%), while WI is rare, with 2 cases (0.86\%). These results suggest that future improvement should focus on primary-diagnosis identification and standardized diagnostic-output formatting.

\begin{table*}[t]
\centering
\small
\setlength{\tabcolsep}{6pt}
\renewcommand{\arraystretch}{1.12}
\begin{tabularx}{0.95\textwidth}{@{}lccrX@{}}
\toprule
\textbf{Error / Discrepancy Type} 
& \textbf{Abbr.} 
& \textbf{Count} 
& \textbf{Percent} 
& \textbf{Description} \\
\midrule
Semantic Equivalence 
& SE 
& 83 
& 35.78\% 
& Uses clinically equivalent synonyms instead of exact reference terms. \\

Overprediction 
& OP 
& 79 
& 34.05\% 
& Provides multiple diagnoses without a clear primary diagnosis, or over-generates secondary conditions. \\

Missed Diagnosis 
& MD 
& 58 
& 25.00\% 
& Fails to identify the core primary diagnosis from the reference answer. \\

Splitting / Combining 
& SD 
& 10 
& 4.31\% 
& Improperly splits a composite diagnosis or combines distinct clinical conditions. \\

Wrong Inference 
& WI 
& 2 
& 0.86\% 
& Identifies partial evidence but draws an incorrect diagnostic conclusion. \\
\bottomrule
\end{tabularx}
\caption{Error and discrepancy categories for EMR Diagnosis outputs. Percentages are computed over 232 categorized discrepancies.}
\label{tab:error_types}
\end{table*}

\subsection{LungKG Construction Details}
\label{app:lungkg_entities}

LungKG is built before KG-constrained reasoning-chain construction through a six-step fusion pipeline:
\begin{itemize}
    \item \textbf{Parsing and standardization:} parse internally constructed pulmonary KG triples and guideline-derived structured sources, normalize them into a shared head--relation--tail schema, map entity types to 15 unified categories, and map relation descriptions into structured relation labels.
    \item \textbf{Entity alignment:} align entities across internally curated pulmonary sources and guideline-derived sources with exact matching, fuzzy matching, and LLM-assisted semantic matching grouped by entity type.
    \item \textbf{Deduplication and fusion:} merge aligned entities, deduplicate triples, preserve source provenance, and assign confidence metadata.
    \item \textbf{Relation completion:} use entity descriptions and LLM-assisted inference to add missing relations for important or sparsely connected entities.
    \item \textbf{Quality checking:} generate graph statistics and run sampled LLM-based validation; additional audit details are kept in supporting documentation.
    \item \textbf{Graph preparation:} prepare the connected LungKG graph for inverse-degree subgraph sampling and evidence serialization.
\end{itemize}

\paragraph{Annotation consistency for LungKG construction.}
\label{app:lungkg_agreement}
For categorical annotation tasks, agreement is often summarized with Kappa statistics~\citep{cohen1968weighted}. For span-based entity and relation annotation, however, treating every unlabeled text span as a negative instance creates a very large and ill-defined negative set. We therefore report pairwise F-agreement, based on precision, recall, and F-score~\citep{powers2020evaluation}, for LungKG entity and relation annotation. Let $A_1$ denote the adjudicated reference annotation set and $A_2$ denote the comparison annotation set. With $N_{\mathrm{match}}$ matched annotations, $N_{A_1}$ reference annotations, and $N_{A_2}$ comparison annotations, precision, recall, and F-agreement are computed as:
\begin{equation}
\small
P=\frac{N_{\mathrm{match}}}{N_{A_2}}, \quad
R=\frac{N_{\mathrm{match}}}{N_{A_1}}, \quad
F=\frac{2PR}{P+R}.
\label{eq:kg_annotation_f}
\end{equation}
For entity annotation, a match is counted only when the entity text span, entity type label, and start/end offsets are identical. For relation annotation, a match is counted only when both argument entities, their spans, the relation label, and the relation direction are identical. Under this protocol, the audited LungKG annotation subset reaches 88.2\% entity F-agreement and 83.8\% relation F-agreement.

\begin{algorithm}[h]
  \small
  \caption{Inverse-Degree Long-Tail Subgraph Sampling}
  \label{alg:longtail_sampling}
  \begin{algorithmic}[1]
    \REQUIRE Graph $G=(V,E)$; maximum size $M$; minimum size $m$; long-tail ratio $\alpha$; number of communities $N$; expansion depth $d$
    \ENSURE Community subgraph list $C$
    \STATE Initialize $C \leftarrow \emptyset$
    \STATE Compute $\deg(v)$ for all $v \in V$
    \STATE Let $\tau$ be the median node degree in $G$
    \STATE $V_{\text{tail}} \leftarrow \{v \in V \mid 0 < \deg(v) \leq \tau\}$
    \STATE $V_{\text{head}} \leftarrow \{v \in V \mid \deg(v) > \tau\}$
    \STATE Sample up to $N\alpha$ anchors without replacement from $V_{\text{tail}}$ with probability $P(v)\propto 1/(\deg(v)+\varepsilon)$
    \STATE Sample $N(1-\alpha)$ anchors uniformly from $V_{\text{head}}$
    \FOR{each $a \in V_{\text{tail}}^{\text{sample}}$}
        \STATE $S \leftarrow \text{BFS}(a,G,\text{depth}=d,\text{max\_size}=M)$
        \IF{$|S| \geq m$}
            \STATE $C \leftarrow C \cup \{\textsc{Community}(S,\textsc{LongTail})\}$
        \ENDIF
    \ENDFOR
    \FOR{each $a \in V_{\text{head}}^{\text{sample}}$}
        \STATE $S \leftarrow \text{BFS}(a,G,\text{depth}=1,\text{max\_size}=M)$
        \IF{$|S| \geq m$}
            \STATE $C \leftarrow C \cup \{\textsc{Community}(S,\textsc{Head})\}$
        \ENDIF
    \ENDFOR
    \STATE \textbf{return} $C$
  \end{algorithmic}
\end{algorithm}

\begin{algorithm}[h]
  \small
  \caption{CoT QA Generation from Subgraphs}
  \label{alg:cot_generation}
  \begin{algorithmic}[1]
    \REQUIRE Community subgraph $C$; LLM
    \ENSURE Parsed QA pairs with optional reasoning chains
    \STATE Serialize nodes and internal edges in $C$ into a graph context string $F$
    \STATE Construct prompt $P$ with graph-faithfulness, safety, reasoning, and structured-output instructions
    \STATE $R \leftarrow \text{LLM}(P,F)$
    \STATE Parse generated QA candidates from $R$
    \FOR{each parsed block $q$}
        \IF{$q$ contains both a question and an answer}
            \STATE Store $q$ using a hash of the question as its identifier
        \ENDIF
    \ENDFOR
    \STATE \textbf{return} parsed QA dictionary
\end{algorithmic}
\end{algorithm}

\subsection{KG-Constrained CoT Construction Details}
\label{app:kg_constrained_cot_details}

This appendix provides pseudocode, generated-candidate statistics, and prompt templates for the KG-constrained CoT construction process used in Lung-R1 training. Figure~\ref{fig:kg_constrained_cot_overview} gives a visual overview of inverse-degree anchor sampling, tail/head community expansion, and graph-grounded CoT QA generation.

\begin{figure*}[t]
  \centering
  \includegraphics[width=1\textwidth]{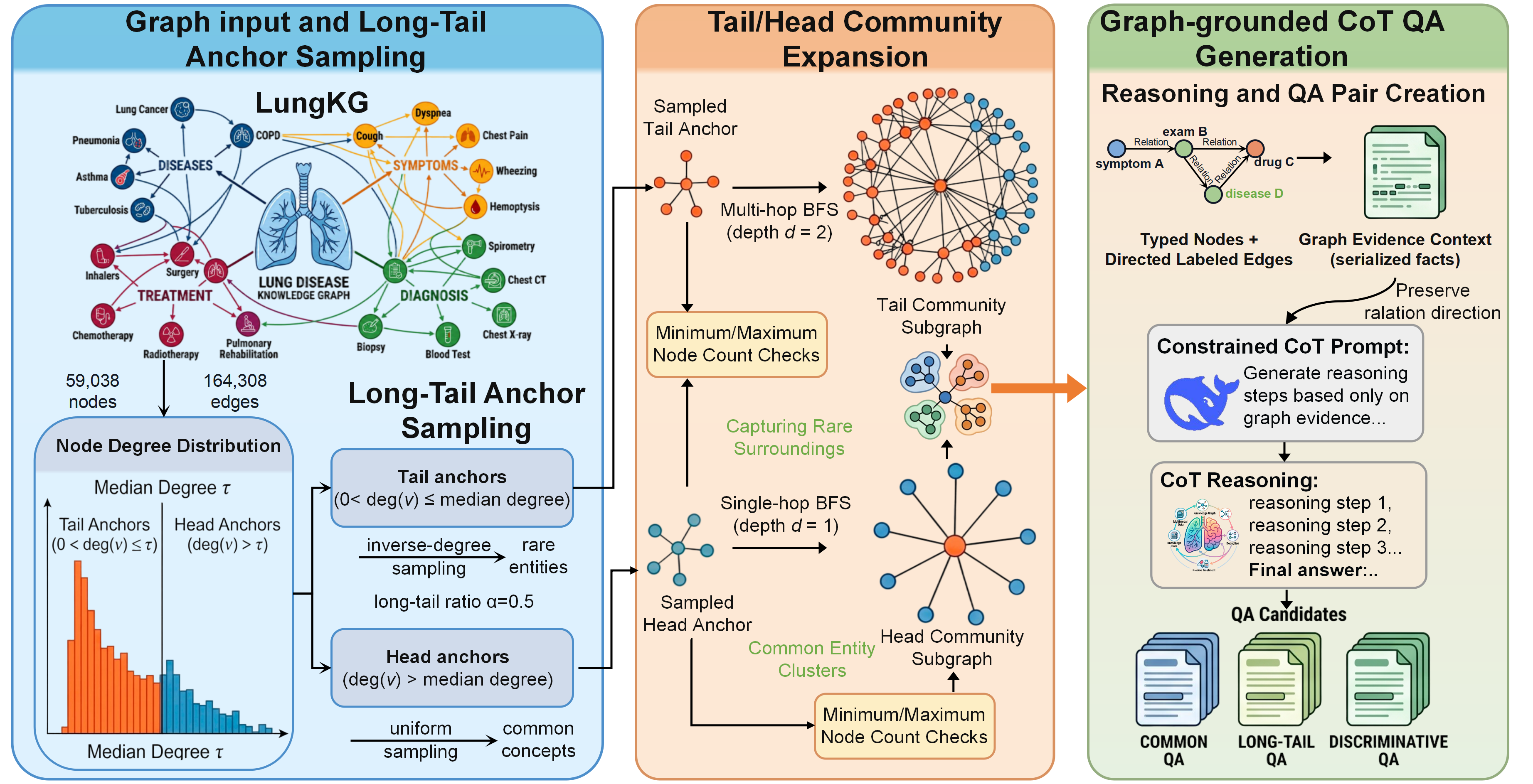}
  \caption{Overview of KG-constrained CoT construction. LungKG subgraphs are sampled with inverse-degree tail-anchor emphasis, expanded into tail/head communities, serialized as directed graph evidence, and converted into graph-grounded QA candidates for downstream filtering.}
  \label{fig:kg_constrained_cot_overview}
\end{figure*}

\paragraph{Inverse-degree sampling.}
Tail anchors are sampled with inverse-degree weights, $P(v)\propto 1/(\deg(v)+\varepsilon)$ for low-degree anchors, while head anchors are sampled uniformly. Algorithm~\ref{alg:longtail_sampling} summarizes the complete sampling procedure, including median-degree tail/head grouping, tail-anchor sampling without replacement, uniform head-anchor sampling, and BFS expansion. We treat this strategy as a long-tail coverage design choice and report the generated distribution descriptively rather than as a causal sampling ablation.

\paragraph{Generated candidate distribution.}
The frozen KG-generated CoT candidate manifest contains 27,571 QA pairs. Table~\ref{tab:kg_cot_distribution} reports the type distribution parsed from this manifest. These counts describe the generated candidate pool before later filtering and conversion into final instruction-tuning files.

\begin{table}[h]
\centering
\small
\begin{tabular}{lcc}
\toprule
\textbf{Type} & \textbf{Count} & \textbf{Proportion} \\
\midrule
Discriminative QA & 10,316 & 37.4\% \\
Long-tail QA & 9,040 & 32.8\% \\
Common QA & 8,215 & 29.8\% \\
\midrule
Total & 27,571 & 100.0\% \\
\bottomrule
\end{tabular}
\caption{Distribution of QA pairs generated by KG-constrained CoT construction in the frozen candidate manifest.}
\label{tab:kg_cot_distribution}
\end{table}

\subsection{Prompt Templates and Evaluation Rubrics}
\label{app:prompt_templates}

This appendix records de-identified prompt templates used by the implementation. Dynamic fields, JSON fields, and sensitive-information placeholders are preserved. These templates do not contain raw EMR text, real graph contexts, credentials, service configuration details, filesystem identifiers, governance identifiers, or review identifiers.

\begin{table*}[t]
  \centering
  \small
  \renewcommand{\arraystretch}{1.12}
  \caption{Prompt template index.}
  \label{tab:prompt_template_index}
  \resizebox{0.96\textwidth}{!}{%
  \begin{tabular}{p{0.18\textwidth}p{0.24\textwidth}p{0.50\textwidth}}
  \toprule
  \textbf{Template} & \textbf{Used for} & \textbf{Inputs and outputs} \\
  \midrule
  KG-CoT QA generation & KG-constrained SFT data construction & Inputs: \texttt{\{num\_qa\}}, \texttt{\{community\_type\_hint\}}, \texttt{\{graph\_context\}}. Outputs: \texttt{type}, \texttt{question}, \texttt{reasoning}, \texttt{answer}, \texttt{evidence}, \texttt{quality}, optional \texttt{distractor}. \\
  KGQA filter & KGQA training-data quality control & Inputs: \texttt{\{input\_text\}}, \texttt{\{output\_text\}}. Output: \texttt{\{"correct": true/false\}}. \\
  EMR filter & EMR diagnosis-label quality control & Inputs: \texttt{\{input\_text\}}, \texttt{\{output\_text\}}. Output: \texttt{\{"correct": true/false\}}. \\
  Pulmonary-QA judge & Pulmonary-QA scoring and QWK & Inputs: \texttt{\{instruction\}}, \texttt{\{input\_text\}}, \texttt{\{ground\_truth\}}, \texttt{\{prediction\}}. Output: \texttt{score}, \texttt{reason}. \\
  EMR Diagnosis judge & EMR Diagnosis scoring and QWK & Inputs: \texttt{\{instruction\}}, \texttt{\{input\_text\}}, \texttt{\{standard\_answer\}}, \texttt{\{model\_answer\}}. Output: \texttt{diagnosis\_score}, \texttt{major\_errors}, \texttt{missing\_key\_points}, \texttt{comments}. \\
  \bottomrule
  \end{tabular}%
  }
\end{table*}

\begin{figure*}[t]
\centering
\begin{tcolorbox}[
  colframe=black!55,
  colback=black!4,
  title={\small KG-CoT QA Generation Prompt},
  fonttitle=\bfseries,
  sharp corners,
  boxrule=0.35mm,
  left=4pt,right=4pt,top=4pt,bottom=4pt
]
\scriptsize
\begin{tabularx}{0.98\textwidth}{p{0.18\textwidth}X}
\textbf{Role} & Act as a rigorous, conservative, and auditable pulmonary medical QA data-generation expert with clinical teaching ability. Output only high-quality samples that are evidence-supported, medically safe, and suitable for SFT. Internally use graph evidence for reasoning, but do not expose graph-retrieval traces in the final natural-language reasoning. \\
\addlinespace[2pt]
\textbf{Inputs} & Use \texttt{\{num\_qa\}}, \texttt{\{community\_type\_hint\}}, and \texttt{\{graph\_context\}}. Generate at most \texttt{\{num\_qa\}} QA samples from the nodes and directed relation edges in the graph context. \\
\addlinespace[2pt]
\textbf{Quality priority} & Medical semantic correctness has the highest priority, followed by graph faithfulness, medical safety, and sample count. If evidence is insufficient, downgrade or abandon the candidate rather than complete it by speculation. \\
\addlinespace[2pt]
\textbf{Graph constraints} & Use only entities and relations explicitly present in the graph context. Preserve relation direction. Do not create extra-graph entities, aliases, relations, mechanisms, treatment plans, examination conclusions, or diagnostic conclusions. Do not rewrite association as causality, prevention as treatment, adverse reaction as indication, or caution as contraindication. Do not infer non-existence from graph absence. \\
\addlinespace[2pt]
\textbf{Medical safety} & Prefer graph evidence about contraindications, cautionary use, severe adverse reactions, and high-risk population restrictions when excluding incorrect conclusions. Do not upgrade possibility to certainty or unsupported alternatives to rankings such as first-line, safer, or best. \\
\addlinespace[2pt]
\textbf{Reasoning} & Common QA must use at least one explicit evidence chain. Long-tail and discriminative QA must use at least two explicit evidence chains and at least three entities. Discriminative QA must support the correct answer and exclude the distractor by explicit evidence. Negative, comparative, or selection-based conclusions must not be built from missing information. \\
\addlinespace[2pt]
\textbf{Output} & Output only structured QA samples. Each sample contains \texttt{type}, \texttt{question}, \texttt{reasoning}, \texttt{answer}, \texttt{evidence}, and \texttt{quality}; discriminative QA must contain \texttt{distractor}. If no qualified sample exists, output an empty result. \\
\end{tabularx}
\end{tcolorbox}
\caption{De-identified KG-constrained CoT QA generation prompt. The template preserves the role, dynamic inputs, graph-faithfulness constraints, safety boundaries, reasoning requirements, and structured output fields used for KG-generated SFT data construction.}
\label{fig:kg_cot_prompt_template}
\end{figure*}

\begin{figure*}[t]
\centering
\begin{tcolorbox}[
  colframe=black!55,
  colback=black!4,
  title={\small KGQA Training-Data Filtering Prompt},
  fonttitle=\bfseries,
  sharp corners,
  boxrule=0.35mm,
  left=4pt,right=4pt,top=4pt,bottom=4pt
]
\scriptsize
\textbf{Use.} Training-data quality control, not final benchmark judging.

\textbf{Inputs.} User message uses \texttt{\{input\_text\}} and \texttt{\{output\_text\}}.

\textbf{Decision rule.} Mark a QA pair as correct only when the final answer accurately answers the question and the reasoning is medically rigorous, with no conceptual confusion, excessive inference, or obvious omission. Mark as incorrect if the answer contains imprecise concepts, contraindication/indication confusion, unsupported causality, dosage-form/ingredient confusion, exaggerated uniqueness, misleading clinical judgment, or inconsistency between reasoning and final answer. If the question lacks sufficient information but the answer gives a specific unsupported conclusion, mark it incorrect.

\textbf{Output.} Output only JSON:
\begin{center}
\texttt{\{"correct": true\}} or \texttt{\{"correct": false\}}.
\end{center}
\end{tcolorbox}
\caption{KGQA training-data filtering prompt. This filter checks whether a generated QA pair is suitable for high-quality fine-tuning data and is separate from the final Pulmonary-QA benchmark judge.}
\label{fig:kgqa_filter_prompt}
\end{figure*}

\begin{figure*}[t]
\centering
\begin{tcolorbox}[
  colframe=black!55,
  colback=black!4,
  title={\small EMR Diagnosis Training-Data Filtering Prompt},
  fonttitle=\bfseries,
  sharp corners,
  boxrule=0.35mm,
  left=4pt,right=4pt,top=4pt,bottom=4pt
]
\scriptsize
\textbf{Use.} Conservative filtering for pulmonary diagnosis labels used in training.

\textbf{Inputs.} User message uses \texttt{\{input\_text\}} for the de-identified record and \texttt{\{output\_text\}} for the diagnosis label.

\textbf{Decision rule.} The main criterion is whether the pulmonary diagnosis clearly contradicts the record. Mark as correct if it is partially supported by symptoms, signs, imaging, laboratory findings, medical history, clinical course, or at least does not conflict with the record. Do not reject an uncertain but non-contradictory label solely because the de-identified snippet lacks every basis for a real clinical diagnosis. Mark as incorrect when the diagnosis conflicts with the record, uses an inconsistent anatomical site or disease type, treats a non-pulmonary diagnosis as pulmonary, or is overly specific with no corresponding evidence.

\textbf{Output.} Output only JSON:
\begin{center}
\texttt{\{"correct": true\}} or \texttt{\{"correct": false\}}.
\end{center}
\end{tcolorbox}
\caption{EMR Diagnosis training-data filtering prompt. The filter accepts uncertain but non-contradictory pulmonary labels and rejects obvious contradictions or unsupported over-specific diagnoses.}
\label{fig:emr_filter_prompt}
\end{figure*}

\begin{figure*}[t]
\centering
\begin{tcolorbox}[
  colframe=black!55,
  colback=black!4,
  title={\small Pulmonary-QA Judge Rubric},
  fonttitle=\bfseries,
  sharp corners,
  boxrule=0.35mm,
  left=4pt,right=4pt,top=4pt,bottom=4pt
]
\scriptsize
\textbf{Use.} Benchmark scoring for respiratory-medicine QA. The same rubric is used by each judge in the five-model ensemble under the fixed decoding protocol in Appendix~\ref{app:evaluation_details}. QWK compares physician-side human scores with rounded ensemble automatic scores on the same model outputs.

\textbf{Inputs.} \texttt{\{instruction\}}, \texttt{\{input\_text\}}, \texttt{\{ground\_truth\}}, \texttt{\{prediction\}}.

\textbf{Scoring scale.}
\begin{itemize}[leftmargin=*,nosep]
  \item \textbf{Score 5:} all core reference points are covered, no medical error, and medically equivalent wording is allowed.
  \item \textbf{Score 4:} mostly correct; about at least 80\% coverage, only minor non-key omissions, and no medical error.
  \item \textbf{Score 3:} partially correct; about 40--70\% coverage, key omissions or vagueness, but no clear medical error.
  \item \textbf{Score 2:} limited coverage or a related but medically non-equivalent concept.
  \item \textbf{Score 1:} serious medical or directional error, or only isolated correct terms.
  \item \textbf{Score 0:} completely wrong, irrelevant, or refusal.
\end{itemize}

\textbf{Procedure and output.} Identify reference core points, check coverage, prioritize medical correctness, and output only JSON: \texttt{\{"score": 0, "reason": "brief judgment rationale"\}}.
\end{tcolorbox}
\caption{Pulmonary-QA judge rubric. The rubric defines the 0--5 ordinal score used by the five-model LLM-as-Judge ensemble for Pulmonary-QA score, score-$\geq4$ rate, and QWK computation.}
\label{fig:pqa_judge_rubric}
\end{figure*}

\begin{figure*}[t]
\centering
\begin{tcolorbox}[
  colframe=black!55,
  colback=black!4,
  title={\small EMR Diagnosis Judge Rubric},
  fonttitle=\bfseries,
  sharp corners,
  boxrule=0.35mm,
  left=4pt,right=4pt,top=4pt,bottom=4pt
]
\scriptsize
\textbf{Use.} Benchmark scoring for model-generated pulmonary diagnoses against reference diagnoses. The prompt is separate from training-data filters and is used by the frozen five-judge ensemble under the fixed decoding protocol.

\textbf{Inputs.} \texttt{\{instruction\}}, \texttt{\{input\_text\}}, \texttt{\{standard\_answer\}}, \texttt{\{model\_answer\}}.

\textbf{Diagnostic correctness scale.}
\begin{itemize}[leftmargin=*,nosep]
  \item \textbf{Score 5:} main diagnoses are all correctly identified with no obvious misdiagnosis.
  \item \textbf{Score 4:} main diagnoses are basically correct, with secondary omissions or wording differences.
  \item \textbf{Score 3:} diagnostic direction is generally correct, but several important diagnoses are missing or mild misdiagnosis exists.
  \item \textbf{Score 2:} clear diagnostic-direction problems with only partial correct clues.
  \item \textbf{Score 1:} most diagnoses are wrong.
  \item \textbf{Score 0:} completely wrong.
\end{itemize}

\textbf{Output.} Output only JSON fields: \texttt{diagnosis\_score}, \texttt{major\_errors}, \texttt{missing\_key\_points}, and \texttt{comments}.
\end{tcolorbox}
\caption{EMR Diagnosis judge rubric. The rubric defines the 0--5 diagnosis score used by the five-model LLM-as-Judge ensemble for EMR Diagnosis score, score-$\geq4$ rate, and QWK computation.}
\label{fig:emr_judge_rubric}
\end{figure*}

\subsection{Lung-R1 Training Hyperparameters}
\label{app:training_config}

Table~\ref{tab:training_config} summarizes the key supervised fine-tuning hyperparameters for the 7B and 14B Lung-R1 variants.

\subsection{KG-Guided Reinforcement Learning Details}
\label{app:kg_guided_rl_details}

This appendix records implementation-level details for the reinforcement-learning stage; the mathematical objective and reward composition are defined in Section~\ref{sec:lung_r1}. After SFT, the model is further optimized with KG-guided RL using prompts sampled from the non-evaluation KGQA and EMR training sources. Evaluation cases are excluded from RL training. The RL stage uses the same diagnosis-oriented response format as SFT: the model first produces a reasoning chain and then gives the final answer or diagnosis.

For each sampled prompt, the policy generates multiple candidate responses under a fixed RL generation configuration. Candidate responses are evaluated by the KG-guided reward function defined in Section~\ref{sec:lung_r1}. The fixed reward weights are $0.50$ for diagnosis correctness, $0.25$ for graph faithfulness, and $0.25$ for relation/path consistency. These weights prioritize the final diagnosis while assigning equal process-level weight to LungKG-supported reasoning and directional relation consistency.

\textbf{Reward scoring details.}
The diagnosis-correctness component uses the same 0--5 diagnosis-alignment rubric used for EMR Diagnosis scoring and normalizes the score to $[0,1]$. The graph-faithfulness component first extracts medical claims from the reasoning chain, then counts whether each claim is fully supported, partially supported, or unsupported by the serialized LungKG evidence. The relation/path-consistency component checks whether extracted relation or path steps preserve valid LungKG relation types and directionality. Unsupported claims are handled through lower graph-faithfulness and relation/path-consistency scores rather than through a separate reward term. All reward components are computed on non-evaluation training prompts; held-out evaluation cases are not used for RL reward construction. This RL stage is intended to refine the SFT checkpoint toward diagnosis-oriented evidence use rather than to introduce new evaluation information.

\subsection{Case Study}
\label{app:case_study}

We selected one de-identified EMR case for qualitative comparison, following the same case-study style as the main text. Figure~\ref{fig:appendix_case_input} shows the input evidence and reference diagnosis. Figures~\ref{fig:appendix_case_lungr1_7b}--\ref{fig:appendix_case_clinicalgpt} show selected model outputs. These examples illustrate how different systems prioritize infection evidence, respiratory-failure severity, and secondary imaging or cardiopulmonary findings; they are not used as a separate statistical evaluation.

\begin{figure*}[t]
\centering
\begin{tcolorbox}[
  colframe=blue!45!black,
  colback=blue!4,
  title={\small EHR Input},
  fonttitle=\bfseries,
  sharp corners,
  boxrule=0.35mm,
  left=5pt,right=5pt,top=5pt,bottom=5pt
]
\scriptsize
\textbf{Chief complaint.}
Fever and shortness of breath for 13 days, worsening with dyspnea for 6 days.

\medskip
\textbf{History of present illness.}
Thirteen days before admission, the patient developed recurrent fever, with a maximum temperature of 39$^\circ$C, accompanied by chills and shortness of breath. Six days later, the patient was transferred to an outside hospital, where sputum smear and culture were performed. After treatment, fever improved, but shortness of breath worsened and the patient developed confusion, requiring tracheal intubation. The patient's condition improved and the ventilator was discontinued with extubation. After extubation, however, shortness of breath and dyspnea worsened again. The patient was re-intubated and fever recurred. The patient was then transferred to the study hospital. Since disease onset, the patient had poor mental status, poor appetite and sleep, reduced physical activity, decreased urine output, normal stool, and no obvious weight change.

\medskip
\textbf{Past medical history.}
The patient had undergone mitral valve replacement more than 20 years earlier and implantable cardioverter-defibrillator implantation seven years earlier, with long-term medication. Type 2 diabetes had been diagnosed half a year earlier and was controlled with medication. The patient had a remote transfusion history and received red blood cells and plasma after onset of the current illness.

\medskip
\textbf{Physical examination and tests.}
The skin and mucosa were mildly yellow, scattered ecchymoses were visible on both upper limbs, mild scleral icterus was present, cardiac dullness was enlarged bilaterally, rhythm was irregular, and breath sounds were decreased in both lower lungs. Chest CT showed left lower-lobe atelectasis, a small amount of subpleural atelectasis in the right lower lobe, possible infection, small bilateral pleural effusions, enlarged cardiac silhouette, and postoperative changes after mitral valve replacement and pacemaker implantation. Laboratory findings included elevated CRP, decreased lymphocyte percentage, and abnormalities in creatinine, GGT, coagulation tests, and D-dimer.

\medskip
\textbf{Ground truth.}
Severe pneumonia.
\end{tcolorbox}
\caption{EHR input for the appendix case. The diagnosis requires integrating fever, inflammatory evidence, respiratory deterioration, intubation/re-intubation, and CT evidence into a severity-aware diagnosis of severe pneumonia.}
\label{fig:appendix_case_input}
\end{figure*}

\begin{figure*}[t]
\centering
\begin{tcolorbox}[
  colframe=green!45!black,
  colback=green!4,
  title={\small Lung-R1-7B Prediction},
  fonttitle=\bfseries,
  sharp corners,
  boxrule=0.35mm,
  left=5pt,right=5pt,top=5pt,bottom=5pt
]
\scriptsize
\textbf{Raw diagnosis output.}
[Severe Pneumonia], [Pulmonary Infection].

\medskip
\textbf{Diagnosis score.}
5 / 5.

\medskip
\textbf{Evaluator comment.}
The answer accurately includes the key reference diagnosis of severe pneumonia. The additional pulmonary-infection label is redundant but does not affect correctness.

\medskip
\textbf{Case-level interpretation.}
This output explicitly preserves the severity-aware target label, severe pneumonia, and then adds pulmonary infection, which is a broader description of the same infectious process. Clinically, the answer prioritizes the acute infectious respiratory-failure syndrome over secondary findings such as atelectasis or pleural effusion.
\end{tcolorbox}
\caption{Lung-R1-7B prediction for the appendix case.}
\label{fig:appendix_case_lungr1_7b}
\end{figure*}

\begin{figure*}[t]
\centering
\begin{tcolorbox}[
  colframe=green!45!black,
  colback=green!4,
  title={\small Lung-R1-14B Prediction},
  fonttitle=\bfseries,
  sharp corners,
  boxrule=0.35mm,
  left=5pt,right=5pt,top=5pt,bottom=5pt
]
\scriptsize
\textbf{Raw diagnosis output.}
[Pulmonary Infection], [Atelectasis], [Respiratory Failure].

\medskip
\textbf{Diagnosis score.}
4 / 5.

\medskip
\textbf{Evaluator comment.}
The answer correctly identifies pulmonary infection and reasonably adds atelectasis and respiratory failure, but it does not explicitly state severe pneumonia.

\medskip
\textbf{Case-level interpretation.}
This output identifies the infectious pulmonary process and the respiratory-failure component, which are both central to the case. However, the raw output does not explicitly state severe pneumonia. It therefore captures the diagnostic direction and major syndrome but under-specifies the final severity-aware diagnosis.
\end{tcolorbox}
\caption{Lung-R1-14B prediction for the appendix case.}
\label{fig:appendix_case_lungr1_14b}
\end{figure*}

\begin{figure*}[t]
\centering
\begin{tcolorbox}[
  colframe=violet!45!black,
  colback=violet!4,
  title={\small Claude-Sonnet-4.5 Prediction},
  fonttitle=\bfseries,
  sharp corners,
  boxrule=0.35mm,
  left=5pt,right=5pt,top=5pt,bottom=5pt
]
\scriptsize
\textbf{Raw diagnosis output.}
Pulmonary infection; left lower-lobe atelectasis; small subpleural atelectasis in the right lower lobe; bilateral pleural effusion; pulmonary hypertension.

\medskip
\textbf{Diagnosis score.}
3 / 5.

\medskip
\textbf{Evaluator comment.}
The diagnostic direction is correct for pulmonary infection, but the answer omits the key severity modifier and places imaging findings and complications beside the main diagnosis, without highlighting the core diagnosis of severe pneumonia.

\medskip
\textbf{Case-level interpretation.}
Claude-Sonnet-4.5 recognizes pulmonary infection but distributes diagnostic weight across several secondary findings. This pattern sees radiological and cardiopulmonary evidence but does not elevate the clinical syndrome to severe pneumonia.
\end{tcolorbox}
\caption{Claude-Sonnet-4.5 prediction for the appendix case.}
\label{fig:appendix_case_claude}
\end{figure*}

\begin{figure*}[t]
\centering
\begin{tcolorbox}[
  colframe=violet!45!black,
  colback=violet!4,
  title={\small GPT-5.1 Prediction},
  fonttitle=\bfseries,
  sharp corners,
  boxrule=0.35mm,
  left=5pt,right=5pt,top=5pt,bottom=5pt
]
\scriptsize
\textbf{Raw diagnosis output.}
[Pulmonary Infection], [Left Lower-Lobe Atelectasis], [Right Lower-Lobe Atelectasis], [Small Bilateral Pleural Effusion], [Pulmonary Hypertension].

\medskip
\textbf{Diagnosis score.}
4 / 5.

\medskip
\textbf{Evaluator comment.}
The answer correctly identifies pulmonary infection and includes objective imaging and complication findings, but it omits the key severity modifier in the reference diagnosis.

\medskip
\textbf{Case-level interpretation.}
GPT-5.1 is evidence-aware but descriptive. It lists infection, atelectasis, effusion, and pulmonary hypertension, yet does not synthesize the critical-care trajectory into a severity-aware diagnosis.
\end{tcolorbox}
\caption{GPT-5.1 prediction for the appendix case.}
\label{fig:appendix_case_gpt51}
\end{figure*}

\begin{figure*}[t]
\centering
\begin{tcolorbox}[
  colframe=orange!60!black,
  colback=orange!5,
  title={\small Qwen2.5-7B-Instruct Prediction},
  fonttitle=\bfseries,
  sharp corners,
  boxrule=0.35mm,
  left=5pt,right=5pt,top=5pt,bottom=5pt
]
\scriptsize
\textbf{Raw diagnosis output.}
[Pulmonary Hypertension], [Small Bilateral Pleural Effusion].

\medskip
\textbf{Diagnosis score.}
2 / 5.

\medskip
\textbf{Evaluator comment.}
The answer only lists imaging and ultrasound findings and does not identify the core diagnosis of severe pneumonia.

\medskip
\textbf{Case-level interpretation.}
Qwen2.5-7B-Instruct misses the infectious diagnosis entirely. It focuses on pulmonary hypertension and pleural effusion, which are secondary or contextual findings rather than the primary diagnosis.
\end{tcolorbox}
\caption{Qwen2.5-7B-Instruct prediction for the appendix case.}
\label{fig:appendix_case_qwen25_7b}
\end{figure*}

\begin{figure*}[t]
\centering
\begin{tcolorbox}[
  colframe=orange!60!black,
  colback=orange!5,
  title={\small ClinicalGPT-R1 Prediction},
  fonttitle=\bfseries,
  sharp corners,
  boxrule=0.35mm,
  left=5pt,right=5pt,top=5pt,bottom=5pt
]
\scriptsize
\textbf{Raw diagnosis output.}
Pulmonary diagnosis: left lower-lobe atelectasis; small subpleural atelectasis in the right lower lobe; possible infection; small bilateral pleural effusion.

\medskip
\textbf{Diagnosis score.}
1 / 5.

\medskip
\textbf{Evaluator comment.}
The answer mainly lists radiological findings and does not form a clear clinical diagnosis. It misses the key reference diagnosis of severe pneumonia and does not reflect disease severity.

\medskip
\textbf{Case-level interpretation.}
ClinicalGPT-R1 mostly reproduces the CT impression and uses the uncertain phrase possible infection. It does not use fever, respiratory failure, intubation, inflammatory markers, and clinical deterioration to produce a firm severe-pneumonia diagnosis.
\end{tcolorbox}
\caption{ClinicalGPT-R1 prediction for the appendix case.}
\label{fig:appendix_case_clinicalgpt}
\end{figure*}

\ifPDFTeX
\end{CJK*}
\fi
\end{document}